\documentclass[letterpaper, 10 pt, conference]{ieeeconf}  

\IEEEoverridecommandlockouts                              

\usepackage{cite}
\usepackage{amsmath,amssymb,amsfonts}
\usepackage{algorithmic}
\usepackage{graphicx}
\usepackage{subcaption}
\usepackage{subfig}
\usepackage{textcomp}
\usepackage{xcolor}
\usepackage{placeins}
\usepackage{booktabs}
\usepackage{multirow}
\usepackage{dsfont}
\usepackage{bm}
\usepackage[export]{adjustbox}
\usepackage{array}
\def\BibTeX{{\rm B\kern-.05em{\sc i\kern-.025em b}\kern-.08em
    T\kern-.1667em\lower.7ex\hbox{E}\kern-.125emX}}
\usepackage{caption}
\begin{document}

\definecolor{poseRed}{rgb}{1.0,0.0,0.0}
\definecolor{poseGreen}{rgb}{0.0,0.8,0.0}
\definecolor{poseBlue}{rgb}{0.0,0.0,1.0}
\definecolor{posePurple}{rgb}{0.6,0.0,0.8}

\title{\LARGE \bf SpotlessGS: Relightable 3D Gaussian Splatting under \\Dynamic Illumination for Robotic Perception
}
\author{
Liang Hong$^{1}$,
Jiaxin Wei$^{1}$,
Simon Schaefer$^{1}$,
Stefan Leutenegger$^{2}$,
Jaehyung Jung$^{1}$
}

\maketitle

\begingroup
\renewcommand\thefootnote{}
\footnotetext{

\copyright\ 2026 IEEE. Personal use of this material is permitted.
Permission from IEEE must be obtained for all other uses, in any current
or future media, including reprinting/republishing this material for
advertising or promotional purposes, creating new collective works,
for resale or redistribution to servers or lists, or reuse of any
copyrighted component of this work in other works.

This work was supported by the EU Horizon Europe project
\mbox{AUTOASSESS}~(Grant No.~101120732).\\
\hspace*{1em}$^{1}$Mobile Robotics Lab, School of Computation, Information and Technology, Technical University of Munich.
E-mail addresses:
\{liang.hong, jiaxin.wei, simon.k.schaefer, jaehyung.jung\}@tum.de \\
\hspace*{1em}$^{2}$Mobile Robotics Lab, Department of Mechanical and Process Engineer-
ing, ETH Zurich.
E-mail address:
stefan.leutenegger@mrl.ethz.ch

}
\addtocounter{footnote}{-1}
\endgroup

\begin{abstract}
Robots operating in dark or poorly lit environments rely on onboard lights, which often produce uneven illumination that degrades downstream perception tasks. Prior approaches based on 2D image enhancement lack reliable supervision and fail to preserve multi-view geometric consistency. To address these limitations, we extend Dark Gaussian Splatting (DarkGS) toward a more accurate and flexible relightable 3D reconstruction framework. First, we eliminate the need for explicit light parameter calibration by jointly optimizing lighting parameters within the Gaussian Splatting framework. Second, we introduce a low-frequency illumination model based on spherical harmonics (SH) to capture spatially varying residual and ambient lighting effects. Third, we incorporate an MLP-based Bidirectional Reflectance Distribution Function (BRDF) to model non-Lambertian reflectance. Experiments on synthetic and real-world datasets demonstrate that our method effectively mitigates illumination artifacts while improving rendering quality and quantitative performance over prior approaches. We further validate its benefits for robotic perception through a downstream task. The code is available at: https://github.com/Liianne/SpotlessGS.
\end{abstract}

\section{Introduction}
\begin{figure}[t]
    \centering

    \begin{subfigure}{0.325\columnwidth}
        \centering
        \small{Robot Platforms}\\[1mm]
        \includegraphics[width=\linewidth]{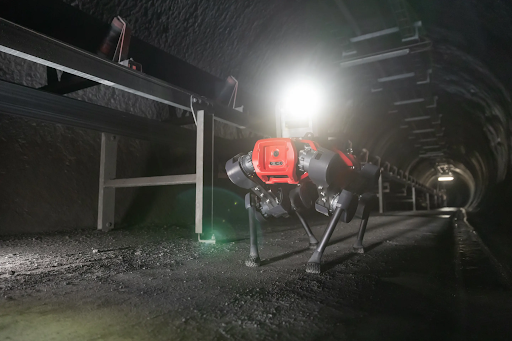}
    \end{subfigure}
    \hfill
    \begin{subfigure}{0.325\columnwidth}
        \centering
        \small{Onboard Images}\\[0.2mm]
        \includegraphics[width=\linewidth]{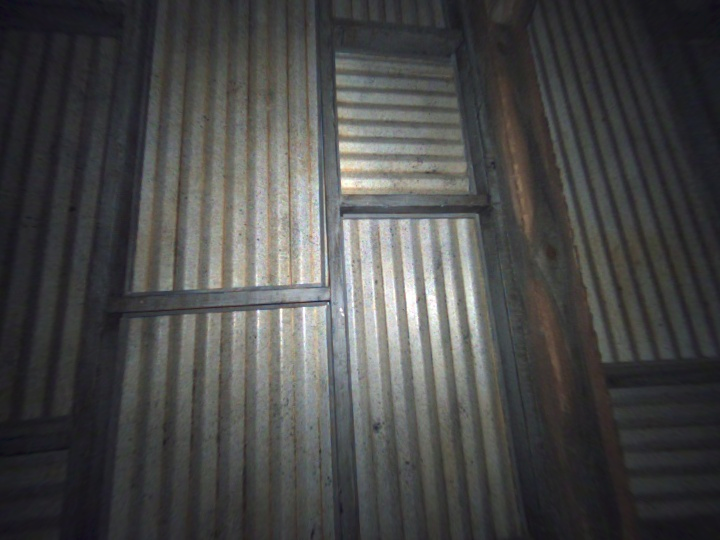}
    \end{subfigure}
    \hfill
    \begin{subfigure}{0.325\columnwidth}
        \centering
        \small{Our Relit Images}\\[0.2mm]
        \includegraphics[width=\linewidth]{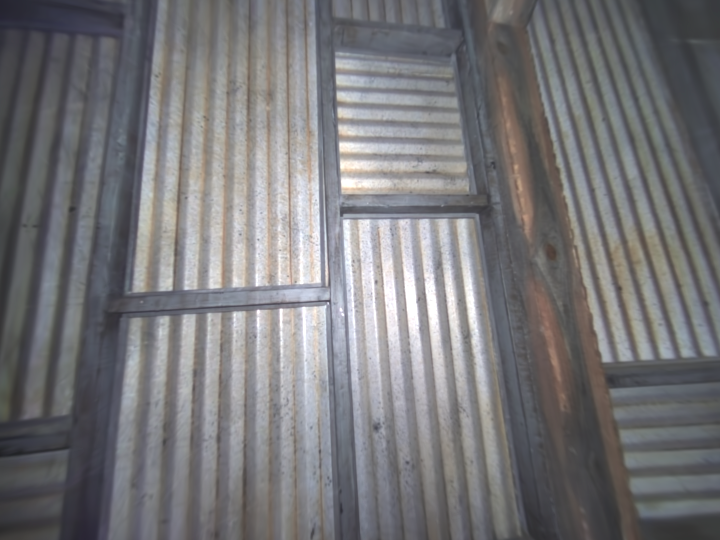}
    \end{subfigure}

    \vspace{1mm}

    \begin{subfigure}{0.325\columnwidth}
        \centering
        \includegraphics[width=\linewidth, trim=0cm 0cm 0cm 0cm,
    clip]{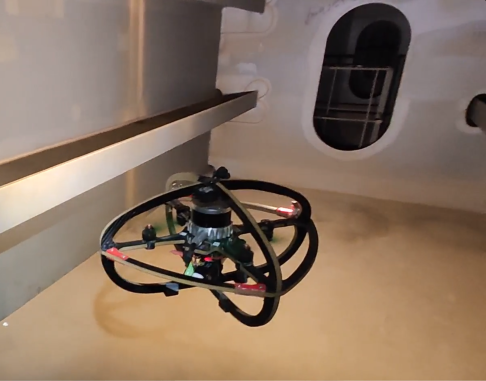}
    \end{subfigure}
    \hfill
    \begin{subfigure}{0.325\columnwidth}
        \centering
        \includegraphics[width=\linewidth]{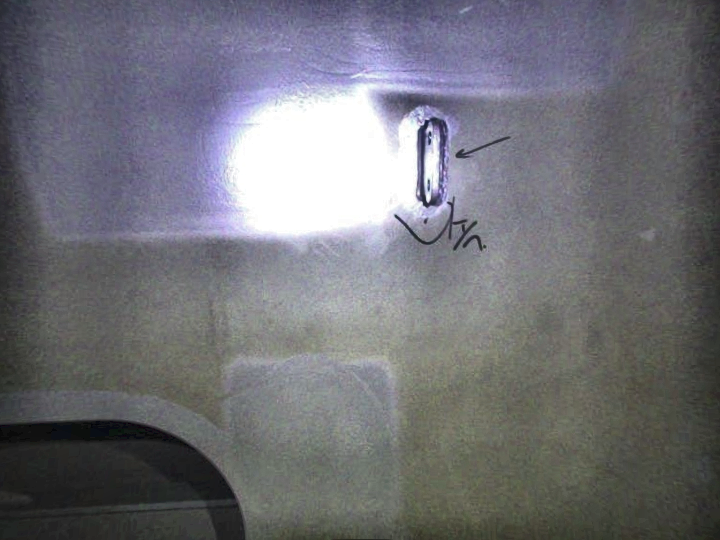}
    \end{subfigure}
    \hfill
    \begin{subfigure}{0.325\columnwidth}
        \centering
        \includegraphics[width=\linewidth]{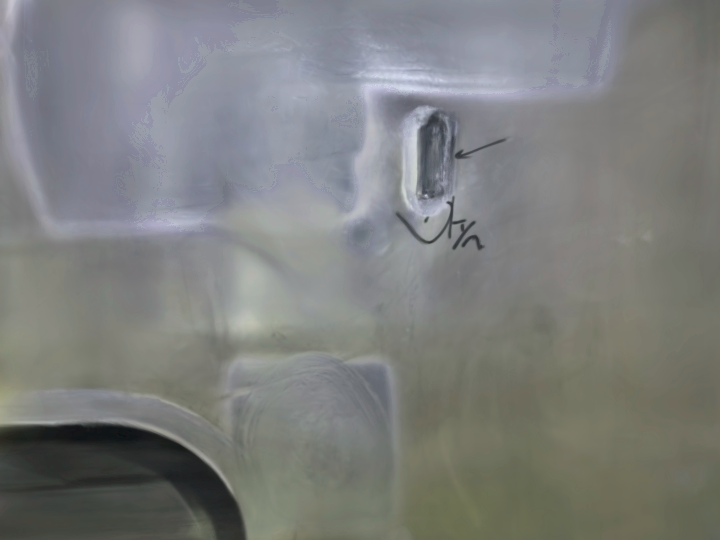}
    \end{subfigure}
    \caption{Robotic platforms and representative onboard images under localized spotlight illumination. Onboard lighting causes over- and under-exposure, producing shadows and saturated highlights that degrade perception. Our method reconstructs a geometrically consistent 3D scene and relights it with uniform illumination, mitigating these artifacts while preserving structure. Robots and onboard images are from~\cite{tranzatto2024team,tranzatto2022cerberus, dharmadhikari2023autonomous}.}
    \label{fig:intro_robots}
\end{figure}

When robots operate in dark environment, an onboard light is commonly employed to light the surroundings. However, such active light source produces uneven illumination, where bright and dark regions coexist due to spotlight effects and surface reflections, as shown in Fig.~\ref{fig:intro_robots}. This results in over- and under-exposed areas that obscure fine textures and degrade downstream perception performance. 

To mitigate these issues, techniques for low-light image enhancement and exposure correction have been proposed~\cite{gao2024relightable, kaleta2025lumigauss}. Most of these methods address globally under- or over-exposed images under the assumption of relatively uniform illumination. However, in real-world robotic scenarios, illumination is often highly non-uniform due to the onboard light source, making these approaches less effective. Moreover, supervised deep learning methods require well exposed ground truth (GT) images~\cite{park2018distort, yu2018deepexposure, cui2022you}, which are difficult to obtain in practice. Although generative models (e.g., Cycle-GAN~\cite{zhu2017unpaired}) relax the need for paired data, purely 2D image-to-image translation methods suffer from limited generalization, poor temporal consistency, and the inability to enforce multi-view geometric constraints, often leading to flickering and inconsistent relighting across viewpoints.

To address these challenges, we explicitly model light sources in 3D space within the Gaussian Splatting framework. By representing the light source directly in the scene, we move beyond appearance-based correction and instead treat illumination as a physically grounded process. This addresses both the lack of GT supervision and the temporal consistency issue: (i) by explicitly modeling the light source, geometric consistency across frames is naturally enforced, eliminating flickering and improving stability; and (ii) by explicitly modeling light sources and relighting the scene in 3D, the method achieves better generalization to real-world environments, as it no longer depends on domain-specific image-to-image mappings.

Among recent advances, DarkGS~\cite{zhang2024darkgs} introduces a relightable 3D reconstruction framework for low-light conditions. While DarkGS models active illumination directly within the Gaussian Splatting representation, it requires a pre-calibration step with a designated calibration target for light parameter estimation and represents global illumination using a single constant scalar term. In addition, its Lambertian reflectance assumption limits its ability to capture complex material dependent light interactions.

Building upon DarkGS, we jointly optimize geometry and illumination parameters in a unified framework, thereby removing a pre-calibration step. To further improve flexibility and visual fidelity, we extend both illumination and reflectance modeling. Specifically, we introduce a spherical harmonics (SH)-based low-frequency illumination model to capture spatially coherent residual and ambient lighting effects beyond the primary spotlight. Furthermore, we replace the Lambertian reflectance with an MLP-based Bidirectional Reflectance Distribution Function (BRDF) to model non-Lambertian material properties. Together, these extensions enable more accurate and expressive simulation of light–geometry interactions under challenging onboard lighting conditions.

Extensive experiments on both synthetic and real-world datasets demonstrate that our method effectively handles uneven illumination, removes spotlight artifacts, produces smoother relighting, and achieves improved temporal consistency compared to prior work.

Our contributions are summarized as follows:
\begin{itemize}
    \item We extend DarkGS by eliminating explicit light parameter calibration and instead jointly optimizing lighting parameters within the Gaussian Splatting framework.
    \item We introduce a SH-based low-frequency illumination model to capture residual and ambient lighting effects beyond the primary spotlight.
    \item We incorporate an MLP-based BRDF to model complex, non-Lambertian reflectance behaviors.
    \item We conduct extensive evaluations on both synthetic and real-world datasets, including data generated from a self-built simulator and a self-collected real-world dataset with well-lit ground truth for quantitative assessment. We further validate our method on in-the-wild datasets and a downstream robotic perception task to demonstrate its practical applicability under challenging lighting conditions.
    
\end{itemize}

\begin{figure*}[t]
    \centering
    \includegraphics[width=\textwidth,
                     trim=0cm 2.5cm 0cm 2.0cm,
                     clip]{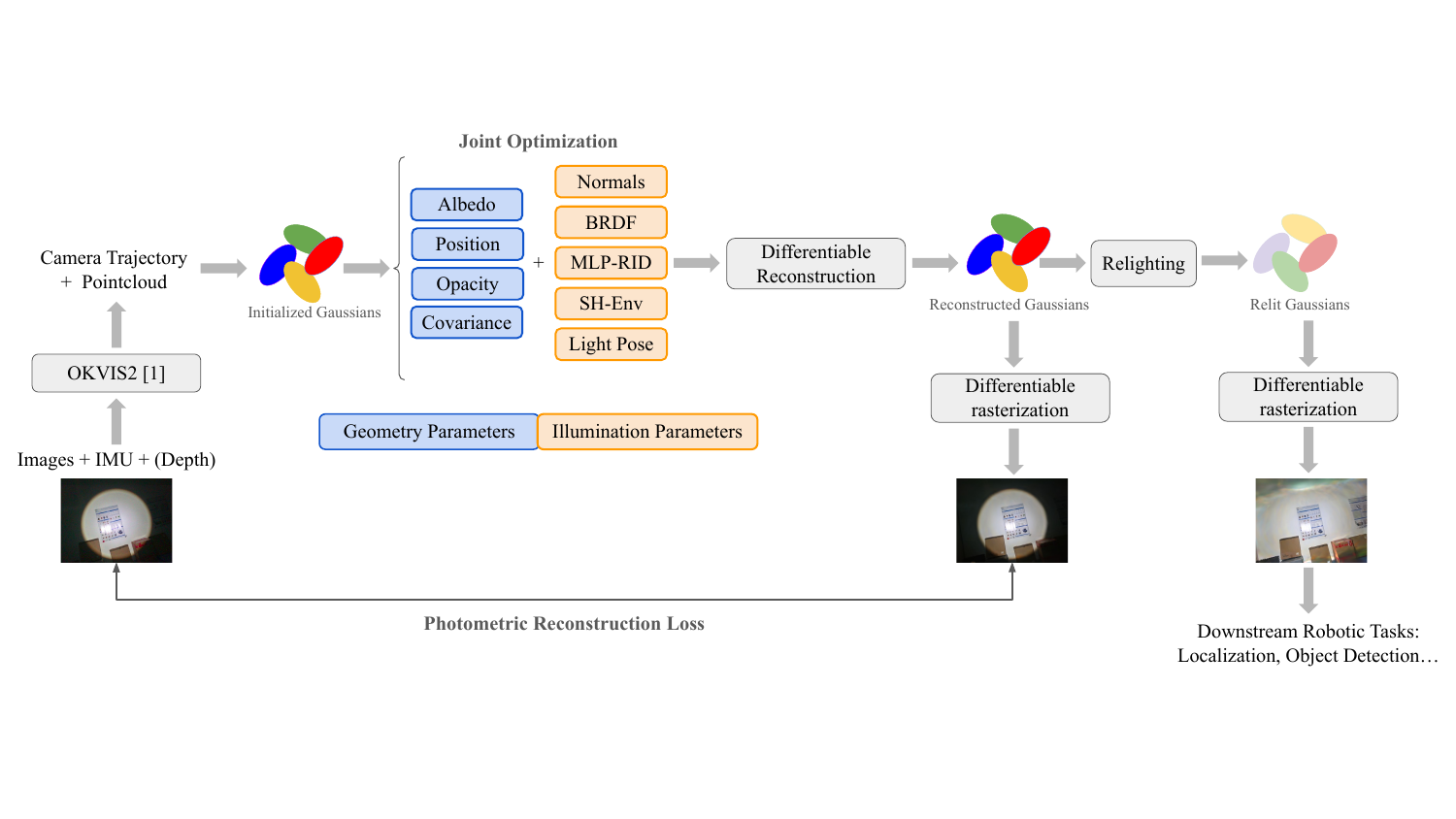}
    \caption{Overview of the SpotlessGS framework.}
    
    \label{fig:overview}
\end{figure*}

\section{Related Work}

\subsection{Image Enhancement}
Traditional image enhancement methods improve images by adjusting contrast, brightness, or tonal distribution. Retinex theory~\cite{land1986alternative} decomposes an image into illumination and reflectance to correct uneven lighting, while histogram equalization~\cite{tomasi1998bilateral} and bilateral filtering~\cite{paris2009bilateral} enhance contrast and preserve edges with high computational efficiency. However, these methods lack semantic understanding and often fail under complex or spatially varying illumination. Learning-based approaches, including CNNs~\cite{park2018distort, yu2018deepexposure} and transformer-based models~\cite{cui2022you}, have achieved strong performance in global exposure correction, but they struggle with uneven illumination and rely on paired ground truth data, which is difficult to obtain in real-world settings. To alleviate the dependency on paired data, generative models such as pix2pix~\cite{isola2017image} and Cycle-GAN~\cite{zhu2017unpaired} have been explored for unpaired image-to-image translation. Although these methods mitigate domain discrepancies in synthetic datasets, their adaptation to real-world scenarios remains limited due to domain gaps and the lack of geometric consistency when operating purely in 2D image space. Therefore, to better address uneven illumination while preserving 3D structural consistency, we shift toward explicitly modeling light sources and performing relighting within a 3D reconstruction framework.

\subsection{Volumetric Relightable Models}
Neural volumetric representations extend NeRF to enable relighting by jointly estimating surface normals, BRDF parameters, and illumination. NeRFactor~\cite{zhang2021nerfactor} recovers geometry together with spatially varying reflectance and illumination, while Ref-NeRF~\cite{verbin2022ref} models view-dependent reflectance to better capture specular effects. These methods achieve high-quality relighting and realistic light–material interactions under novel lighting conditions. Despite their strong rendering performance, volumetric neural fields are computationally expensive and memory-intensive, typically requiring dense multi-view supervision and long training times. Moreover, their heavy optimization and inference costs make them difficult to deploy in real-time robotic systems, where efficiency and robustness under dynamic lighting are critical.

\subsection{Point-base Relightable Models}
Relightable 3D Gaussian~\cite{gao2024relightable} extends standard Gaussian Splatting (GS) by incorporating surface normals, BRDF parameters, and directional lighting, enabling differentiable physically based rendering under varying illumination. However, their performance depends on clean inputs and consistent lighting, limiting robustness in uncontrolled real-world environments. LumiGauss~\cite{kaleta2025lumigauss} further decouples lighting and geometry from unconstrained photo collections to support realistic relighting under novel environment maps, but performance may degrade under low-light or highly uneven illumination. To address such scenarios, DarkGS~\cite{zhang2024darkgs} explicitly models light sources and jointly optimizes geometry, reflectance, and lighting in GS framework. Nevertheless, it adopts simplifying assumptions, including constant ambient light and Lambertian reflectance. Moreover, it depends on a pre-calibration pipeline to initialize illumination parameters, which may not be practical in real-world robotic deployments where hardware configurations are unavailable. Building upon DarkGS, we enhance ambient illumination using spherical harmonics, model reflectance with an MLP-based BRDF, and jointly optimize lighting and geometry without a pre-calibration step. 

\section{Methodology}
\subsection{Overview}

Fig.~\ref{fig:overview} presents the overall pipeline. Given multi-view RGB images and IMU data, optionally with associated depth images, we first estimate camera poses and a sparse point cloud using OKVIS2~\cite{leutenegger2022okvis2}. Then, the camera poses and point cloud are used to initialize a set of 3D Gaussians, where each Gaussian is associated with geometry and illumination parameters. Specifically, the Gaussian parameters are defined as $\mathcal{G}=\{\mathbf{p}_i,\mathbf{\Sigma}_i,\alpha_i,\mathbf{a}_i,\mathbf{n}_i,\rho_i,\eta_i\}_{i=1}^{N}$, where $\mathbf{p}_i$ denotes the 3D position, $\mathbf{\Sigma}_i$ is the covariance, $\alpha_i$ is the opacity, $\mathbf{a}_i$ is the albedo, $\mathbf{n}_i$ is the surface normal, $\rho_i, \eta_i$ are the roughness and metalness parameters, and $N$ is the total number of 3D Gaussians used to represent the scene. The illumination parameters are defined as $\mathcal{L}=\{\mathbf{c},\theta_{\Phi},\theta_f,\mathbf{T}\}$, where $\mathbf{c}$ are the spherical harmonics coefficients, $\theta_{\Phi}$ are the parameters of the incident light MLP, $\theta_f$ are the parameters of the BRDF MLP, and $\mathbf{T}=(\mathbf{R},\mathbf{t})$ denotes the relative pose between  a spotlight source and a camera. The complete set of optimized parameters is therefore given by $\Theta=\{\mathcal{G},\mathcal{L}\}$, which are jointly optimized under the photometric reconstruction loss within a differentiable rendering framework. Finally, we relight the scene by replacing the learned incident with a constant value to have a more even lighting environment. The relit images can be used for downstream robotic perception tasks, such as localization and object-level scene understanding.

\subsection{Light Model}
\subsubsection{Incident Light Model}

We model the spotlight as the primary directional light source, corresponding to the onboard illumination commonly used in robotic platforms. Following DarkGS~\cite{zhang2024darkgs}, we adopt a neural Radiant Intensity Distribution (RID) to model the angular fall-off and spatial distribution of the light. The RID is parameterized by an MLP, enabling flexible approximation of different spotlight patterns.

Specifically, let $\mathbf{x}$ denote a 3D point in the scene and 
$\mathbf{o}$ the light source position. 
The direction from the light to the point is defined as 
$\boldsymbol{\omega}_\mathbf{x} = \mathbf{x} - \mathbf{o}$, 
and $\boldsymbol{\omega}_l$ denotes the central axis of the spotlight. 
The incident light intensity at $\mathbf{x}$ is modeled as a function 
of the angle between $\boldsymbol{\omega}_\mathbf{x}$ and $\boldsymbol{\omega}_\mathbf{l}$~\cite{zhang2024darkgs}:
\begin{equation}
\Phi_{\theta_{\Phi}}(\mathbf{x}) 
= \mathrm{MLP}_{\theta_{\Phi}}\!\left(
\cos^{-1}\!\left(
\frac{\bm{\omega}_\mathbf{x}}{\|\bm{\omega}_\mathbf{x}\|_2}
\cdot
\bm{\omega}_\mathbf{l}
\right)
\right),
\end{equation}
where $\theta_{\Phi}$ denotes the learnable parameters of the MLP. 
This formulation provides a compact and expressive representation 
of directional light distributions suitable for relightable 3D 
reconstruction under onboard illumination.

\subsubsection{Low-frequency Illumination Model}

In real robotic environments, onboard spotlights often introduce additional low-frequency illumination effects, such as scattering and indirect bounce light, which cannot be fully explained by a single spotlight model. Moreover, the scene may also contain ambient illumination components. Modeling these effects using a constant ambient term, as in DarkGS, is insufficient to capture spatially coherent lighting variations. To address this limitation, we introduce a spherical harmonics parameterized low-frequency illumination model. The SH basis provides a compact and expressive representation of smooth angular illumination variations. Specifically, the low-frequency illumination at a surface point $\mathbf{x}$ is modeled as
\begin{equation}
    L_{\mathrm{lf}}(\mathbf{x}) = \mathbf{c}^{\top} \mathbf{Y}\big(\mathbf{n}(\mathbf{x}), \mathbf{l}(\mathbf{x})\big),
\end{equation}
where $\mathbf{n}(\mathbf{x})$ denotes the surface normal, $\mathbf{l}(\mathbf{x})$ is the light direction, $\mathbf{Y}(\cdot)$ represents the real spherical harmonics basis functions, and $\mathbf{c} \in \mathbb{R}^{16}$ are learned SH coefficients.

The overall incident radiance at point $\mathbf{x}$ is then given by
\begin{equation}
    L_i(\mathbf{x}) = \Psi_{\tau}(\mathbf{x})\,\Phi_{\theta_{\Phi}}(\mathbf{x}) + L_{\mathrm{lf}}(\mathbf{x}),
\end{equation}
where $\Phi_{\theta_{\Phi}}(\mathbf{x})$ models the incident light (primary spotlight emission), $\Psi_{\tau}(\mathbf{x})$ is the light fall-off function~\cite{zhang2024darkgs}, and $L_{\mathrm{lf}}(\mathbf{x})$ captures ambient and other low-frequency residual illumination effects. This formulation generalizes traditional ambient lighting models while improving robustness under complex onboard lighting conditions.

\subsubsection{BRDF Model}
In DarkGS, the BRDF is simplified to a Lambertian model, which assumes purely diffuse reflection and ignores material-dependent effects such as specularity, roughness, or metallicity. To approximate a broad range of reflectance behaviors beyond the Lambertian assumption, we replace the Lambertian formulation with a learnable BRDF module parameterized by a lightweight MLP. We use an MLP whose weights are shared across all Gaussians splats. Formally, the reflectance at a surface point is modeled as:
\begin{equation}
f_r(\mathbf{v}, \mathbf{l}, \mathbf{n}, \rho, \eta)
= \mathrm{MLP}_{\theta_f}\Big( \gamma(\mathbf{v}), \gamma(\mathbf{l}), \gamma(\mathbf{n}), \rho, \eta \Big),
\end{equation}
where $\mathbf{v}$, $\mathbf{l}$, and $\mathbf{n}$ are the view direction, light direction, and surface normal, $\gamma(\cdot)$ denotes sinusoidal positional encoding with multiple frequency bands, $\rho$ is roughness, $\eta$ is metalness, and $\theta_f$ is the learnable parameters of the MLP.



\subsection{Relightable Gaussian Splatting Framework}
We extend Gaussian Splatting to enable physically consistent relighting by augmenting each anisotropic Gaussian with additional surface and illumination attributes. In addition to the standard geometric parameters, each Gaussian is associated with a surface normal and learnable BRDF parameters (e.g., roughness and metalness, initialized to 0.5 and 0.0, respectively). These attributes are processed by a lightweight MLP to model spatially varying reflectance under different viewpoints and illumination conditions.


\subsubsection{Rendering Equation}
We extend the rendering equation of DarkGS by incorporating a learnable BRDF, a spotlight model, and ambient lighting via spherical harmonics. The radiance rendered at pixel $(u,v)$ is computed as:
\begin{equation}
\label{eq:rendering}
L_{u,v} =
\sum_{i=1}^{N}
\rho_i f_r(\boldsymbol{\omega}_i,\mathbf{l}_i,\mathbf{n}_i)
(L_\mathrm{spot}+L_{\mathrm{lf}})\alpha_i
\prod_{j<i}(1-\alpha_j),
\end{equation}
where $\rho_i$ is the learnable albedo, $f_r$ is an MLP-based BRDF, $L_\mathrm{spot}$ denotes the direct illumination from the spotlight, modeled as $\Psi_{\tau}(\mathbf{x})\,\Phi_{\theta_{\Phi}}(\mathbf{x})$, $L_{\mathrm{lf}}$ represents the low-frequency ambient component using spherical harmonics, and $N$ is the total number of 3D Gaussians. We construct the rendered image \( \hat{I} \) by evaluating Eq.~(\ref{eq:rendering}) at each pixel.

\subsubsection{Training Loss Function}
To supervise the reconstructed output, we adopt a photometric loss that combines pixel-wise L1 loss and the Structural Similarity Index (SSIM). Additionally, we regularize the estimated light-to-camera pose to improve physical plausibility and stabilize optimization in the early training phase. The total training loss is defined as:

\begin{equation}
\begin{aligned}
\mathcal{L}_{\text{total}}
&= (1-\lambda_{\text{SSIM}})\|\hat{I}-I\|_1
+ \lambda_{\text{SSIM}}\big(1-\mathrm{SSIM}(\hat{I}, I)\big) \\
&\quad + \mathds{1}_{\mathrm{calib}}
\,\lambda_{\text{trans}}
\left\|\mathbf{t}_{\text{light}}-\mathbf{t}_{\text{prior}}\right\|_2,
\end{aligned}
\end{equation}
where \( \hat{I} \) is the reconstructed image, \( I \) is the input image captured under onboard illumination, and \( \text{SSIM}(\hat{I}, I) \) denotes the structural similarity index. The parameter \( \lambda_{\text{SSIM}} \in [0,1] \) balances pixel accuracy and perceptual quality. The last term regularizes the predicted light translation \( \mathbf{t}_{\text{light}} \) toward a prior \( \mathbf{t}_{\text{prior}} \), which can be obtained during data collection through manual measurement (e.g., a physical ruler). \(\mathds{1}_{\mathrm{calib}}\) denotes a binary indicator that equals 1 when a light translation prior is available and 0 otherwise, enabling the regularization term only when prior information is provided. Weighted by \( \lambda_{\text{trans}} \), this regularization discourages degenerate light configurations while allowing the lighting parameters to be learned directly from image observations when no prior is available.

\subsubsection{Relighting Pipeline}
To enable relighting without the need to learn additional light-related parameters, we replace the learned spotlight intensity function \( L_\mathrm{spot} \) with a fixed constant vector \( L_0 \). This effectively disables the spotlight effect and instead simulates a uniform illumination across the scene, similar to an environment light with constant intensity. The corresponding rendering equation becomes:
\begin{equation}
\hat{L}_{u,v} = \sum_{i \in N} \rho_i f_r(\boldsymbol{\omega}_i, \mathbf{l}_i, \mathbf{n}_i) (L_0 + L_{\mathrm{lf}}) \alpha_i \prod_{j<i} (1 - \alpha_j),
\end{equation}
where \( L_0 \in \mathbb{R}^3 \) is a constant RGB illumination vector, and \( L_{\mathrm{lf}} \) represents the ambient light component estimated by spherical harmonics. This configuration allows for relighting under generic lighting conditions without re-training or fine-tuning the light model, making it suitable for applications where only geometry and material estimation are available or when real-time relighting is required.

\section{Experiment}
\subsection{Experimental Setup}

\subsubsection{Synthetic Datasets}
\begin{figure}[t]
\centering

\begin{subfigure}{\columnwidth}
  \centering
  \includegraphics[width=\columnwidth, trim=0cm 0cm 0cm 0.0cm, clip]{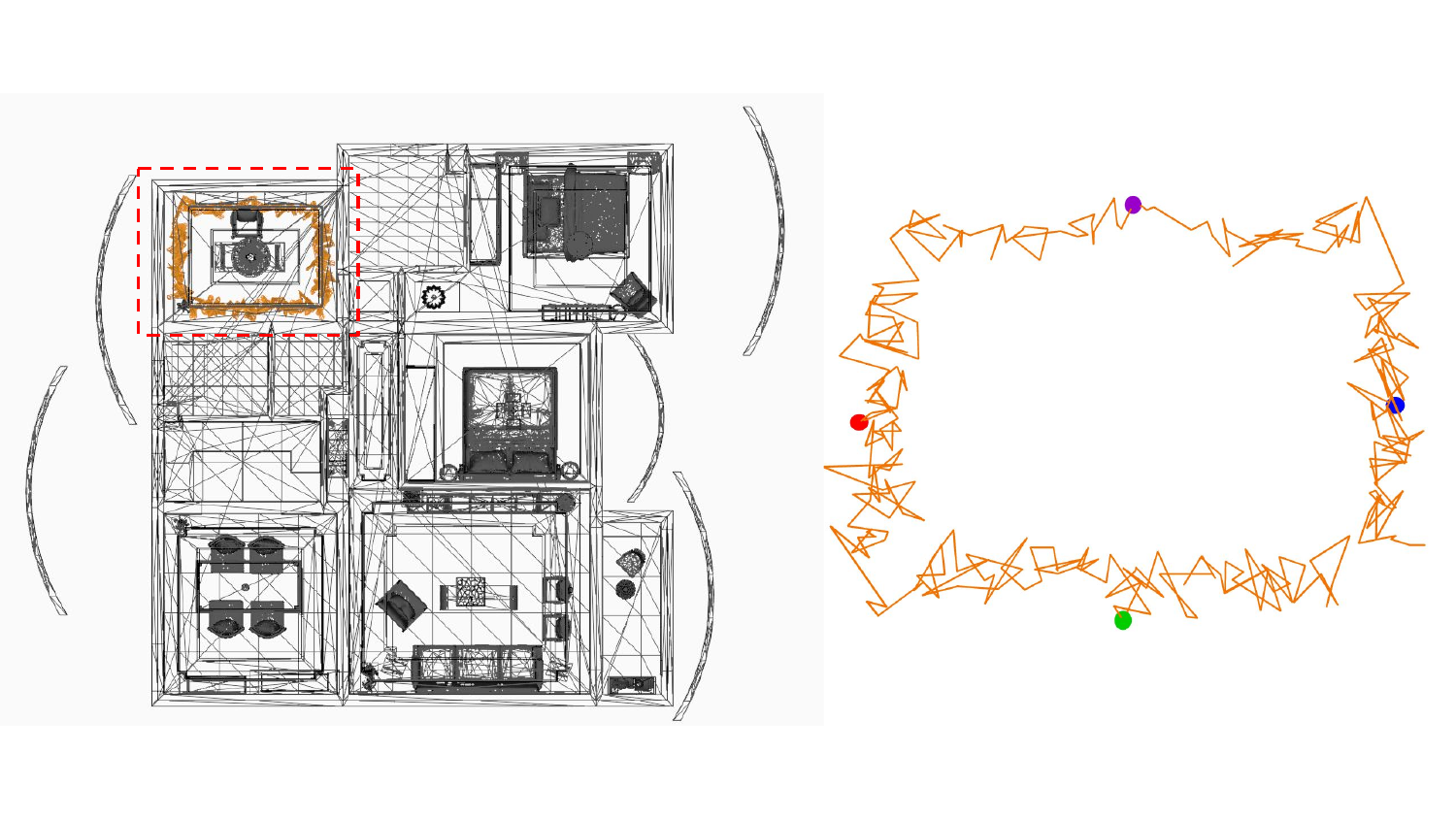}
  \caption{Sample camera trajectory (orange) in a 3D-FRONT scene. Colored markers indicate the four sampled poses.}
  \label{fig:sim_traj}
\end{subfigure}

\vspace{1mm}

\begin{subfigure}{\columnwidth}
  \centering

  \begin{subfigure}{0.243\columnwidth}
    \centering
    \footnotesize Pose 1 \textcolor{poseRed}{\textbullet}\\[1mm]
    \includegraphics[width=\linewidth]{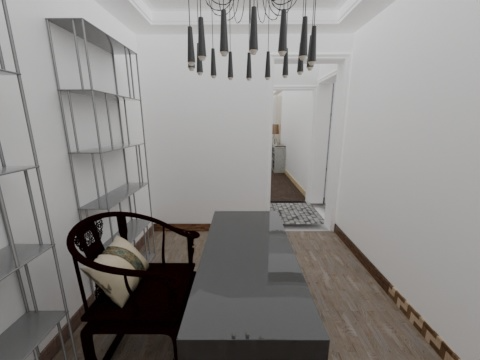}
  \end{subfigure}\hfill
  \begin{subfigure}{0.243\columnwidth}
    \centering
    \footnotesize Pose 2 \textcolor{poseGreen}{\textbullet}\\[1mm]
    \includegraphics[width=\linewidth]{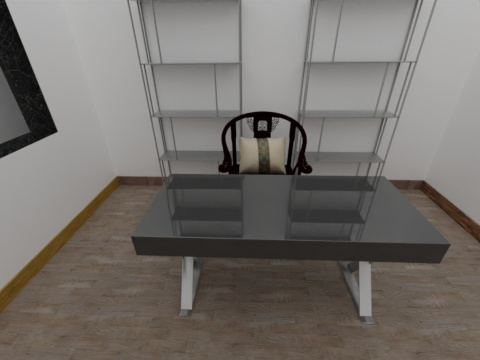}
  \end{subfigure}\hfill
  \begin{subfigure}{0.243\columnwidth}
    \centering
    \footnotesize Pose 3 \textcolor{poseBlue}{\textbullet}\\[1mm]
    \includegraphics[width=\linewidth]{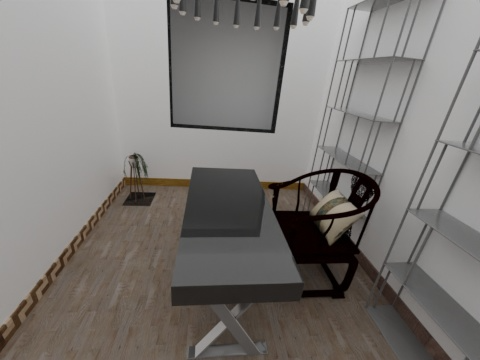}
  \end{subfigure}\hfill
  \begin{subfigure}{0.243\columnwidth}
    \centering
    \footnotesize Pose 4 \textcolor{posePurple}{\textbullet}\\[1mm]
    \includegraphics[width=\linewidth]{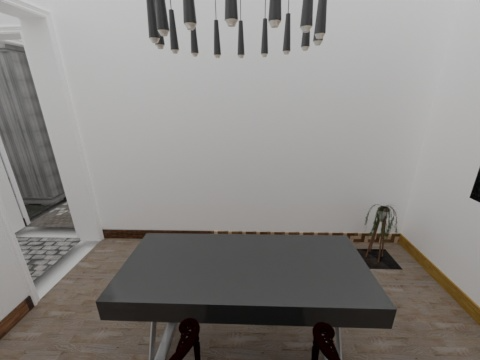}
  \end{subfigure}

  \vspace{0.5mm}

  \begin{subfigure}{0.243\columnwidth}
    \centering
    \includegraphics[width=\linewidth]{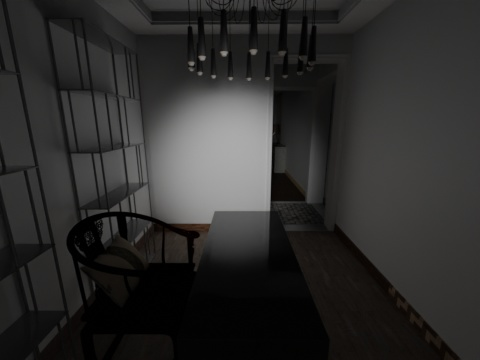}
  \end{subfigure}\hfill
  \begin{subfigure}{0.243\columnwidth}
    \centering
    \includegraphics[width=\linewidth]{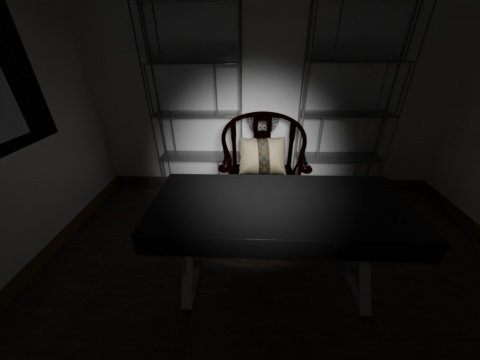}
  \end{subfigure}\hfill
  \begin{subfigure}{0.243\columnwidth}
    \centering
    \includegraphics[width=\linewidth]{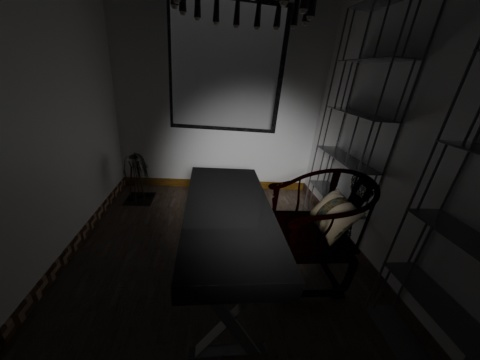}
  \end{subfigure}\hfill
  \begin{subfigure}{0.243\columnwidth}
    \centering
    \includegraphics[width=\linewidth]{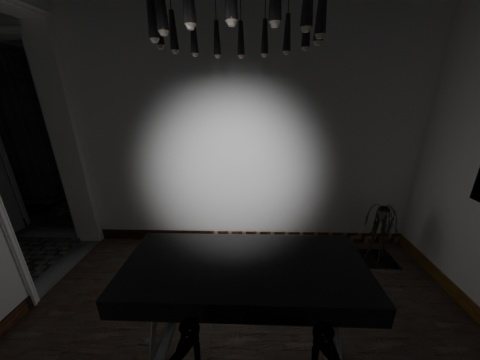}
  \end{subfigure}

  \caption{Paired observations at the four sampled poses: top row shows uniformly illuminated ground truth; bottom row shows uneven illumination under the simulated onboard spotlight.}
  \label{fig:sim_pairs}
\end{subfigure}

\caption{Photo-realistic simulator in Blender for generating synthetic dataset.}
\label{fig:simulator}
\end{figure}

We develop a simulator in Blender to generate paired unevenly illuminated and uniformly illuminated images for controlled evaluation. Scenes are constructed using 3D-FRONT~\cite{fu20213d}. To simulate onboard illumination, we configure a spotlight that emits a cone-shaped beam along a specified direction. For uneven lighting, only the spotlight is active while all other light sources are disabled. For ground truth images, the spotlight is turned off and all lamps and ceiling lights are activated to produce a uniformly illuminated environment. Camera poses are sampled along a rectangular trajectory enclosing the room interior in the $x$–$y$ plane, with a fixed height in the $z$ direction. Positions are uniformly sampled along the trajectory, and camera orientations are aligned perpendicular to the path direction. To simulate realistic motion, Gaussian noise (mean of $0\,\textrm{m}$, standard deviation of $0.1\,\textrm{m}$ for position and $0.1\,\textrm{rad}$ for rotation) is applied. Fig.~\ref{fig:simulator} shows an example of generated camera trajectory and four sampled paired images along the trajectory.


\subsubsection{Real-world Datasets}
We further evaluate our method on real-world datasets. First, we evaluate our method on the DarkGS dataset~\cite{zhang2024darkgs} to ensure fair comparison. Then, to enable quantitative evaluation, we also collect a real-world dataset with an onboard lighting setup, where corresponding uniformly illuminated images are captured as GT. Finally, to validate our approach under more challenging and realistic conditions, we conduct experiments on two publicly available real-world datasets: the ANYmal Dataset~\cite{tranzatto2024team, tranzatto2022cerberus} and the Vessel Dataset~\cite{dharmadhikari2023autonomous}. 


\subsubsection{Baselines}
We compare our method with Vanilla 3DGS~\cite{kerbl20233d}, which assumes static illumination, and DarkGS~\cite{zhang2024darkgs}, which is specifically designed for low-light environments with an onboard spotlight. This comparison enables a comprehensive evaluation of the advantages of our framework over both the original formulation and illumination-aware extensions. For fair comparison with DarkGS, we use the official DarkGS implementation and follow their default training protocol. On the DarkGS dataset provided in the original paper, we train the model with the same configuration as described in~\cite{zhang2024darkgs}. On our real-world dataset, where explicit light calibration is impractical, we remove the pre-calibration step and instead optimize lighting parameters jointly within the training process. To avoid disadvantaging DarkGS due to suboptimal lighting initialization without light pre-calibration, we additionally initialize DarkGS using the optimized light parameters estimated by our method, treating them as pseudo pre-calibrated values. We report the best performance of these two settings achieved by DarkGS for evaluation.

\subsubsection{Evaluation Metrics}
\label{sec:eval-metric}
On the synthetic and self-collected real-world dataset, where we have access to GT relit images, we evaluate Peak Signal-to-Noise Ratio (PSNR), SSIM and Learned Perceptual Image Patch Similarity (LPIPS)~\cite{zhang2018perceptual}. To evaluate the consistency across consecutive frames, we also evaluate temporal PSNR and SSIM computed from a pair of consecutive relit images assuming view differences are small to study the effect of multi-view rendering.

\subsubsection{Dataset Splitting}
For evaluation, we adopt a view-splitting protocol. On the synthetic, DarkGS, ANYmal, and Vessel datasets, every 11th image is selected as a test sample, while the remaining images are used for training. For the self-collected real-world dataset, a separate segment of the trajectory captured under uniform illumination is reserved for evaluation. As a result, the synthetic dataset contains 287 training images, 29 testing images. The self-collected real-world dataset consists of 774 training images and 61 testing images.

\subsection{Results}
\subsubsection{Relighting Evaluation}
\begin{figure*}[!tbp]
\centering

\setlength{\tabcolsep}{0.5mm} 

\newcommand{\imgw}{0.18\linewidth}
\newcommand{\labw}{1.3em}

\newcommand{\RowLabel}[1]{%
  \adjustbox{valign=c}{\rotatebox{90}{\small #1}}}

\newcommand{\Img}[1]{%
  \adjustbox{valign=c}{\includegraphics[width=\imgw]{#1}}}

\begin{tabular}{
@{} m{\labw} @{\hspace{1mm}}
>{\centering\arraybackslash}m{\imgw}
>{\centering\arraybackslash}m{\imgw}
>{\centering\arraybackslash}m{\imgw}
>{\centering\arraybackslash}m{\imgw}
>{\centering\arraybackslash}m{\imgw}
@{}}
& {\small Input}
& {\small Vanilla 3DGS}
& {\small DarkGS}
& {\small Ours}
& {\small GT}
\end{tabular}

\vspace{1mm} 

\begin{tabular}{
@{} >{\centering\arraybackslash}m{\labw} @{\hspace{1mm}}
c c c c c @{}
}

\RowLabel{Uneven} &
\Img{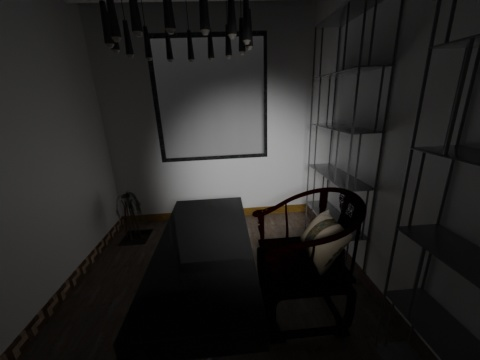} &
\Img{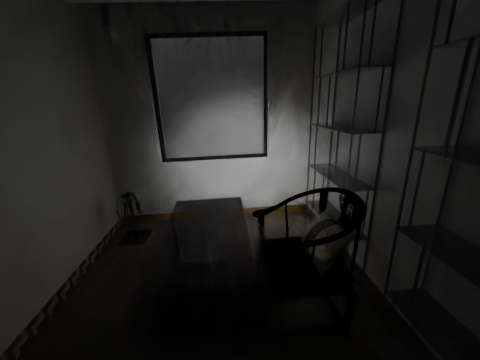} &
\Img{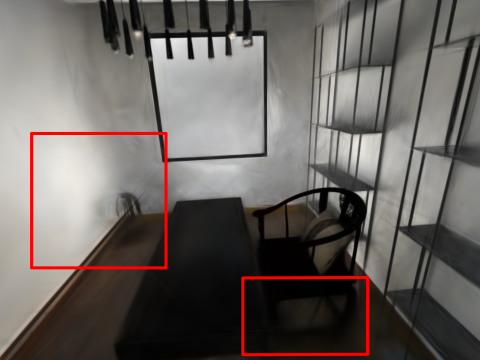} &
\Img{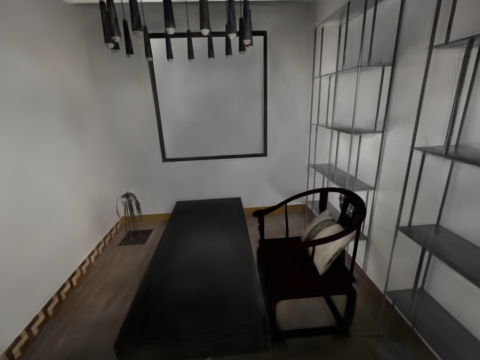} &
\Img{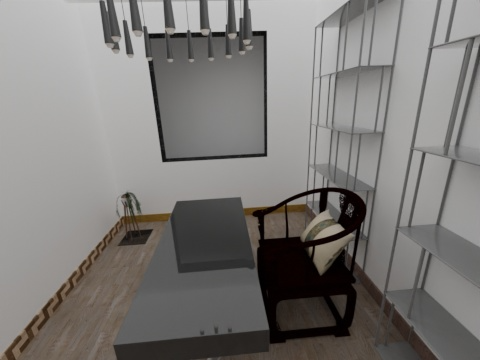} \\
\noalign{\vspace{1mm}}

\RowLabel{Saturated} &
\Img{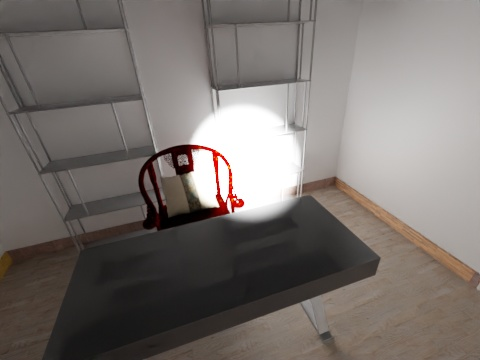} &
\Img{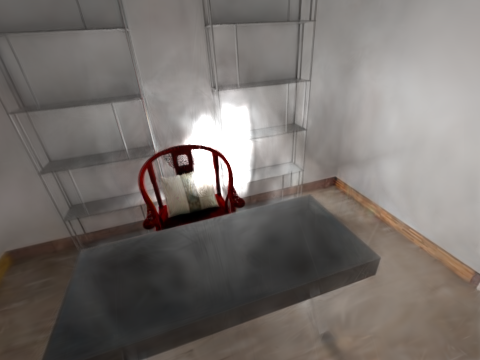} &
\Img{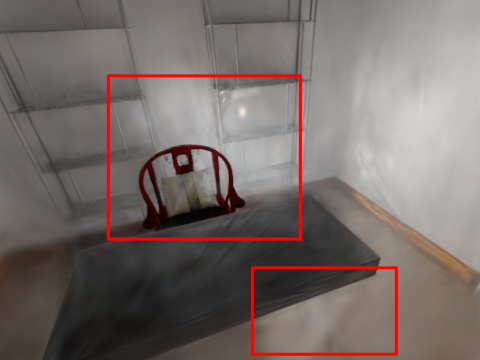} &
\Img{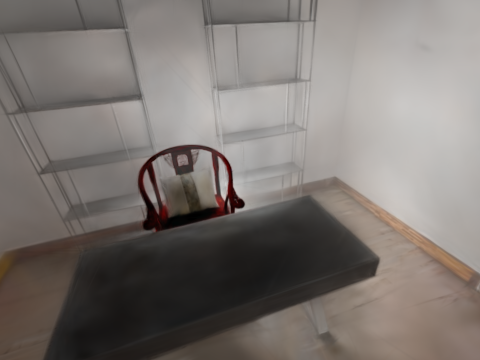} &
\Img{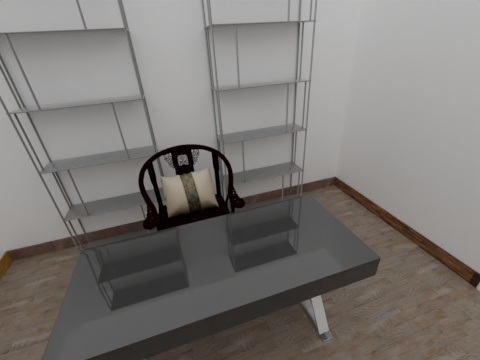} \\
\noalign{\vspace{1mm}}

\RowLabel{Self-col.} &
\Img{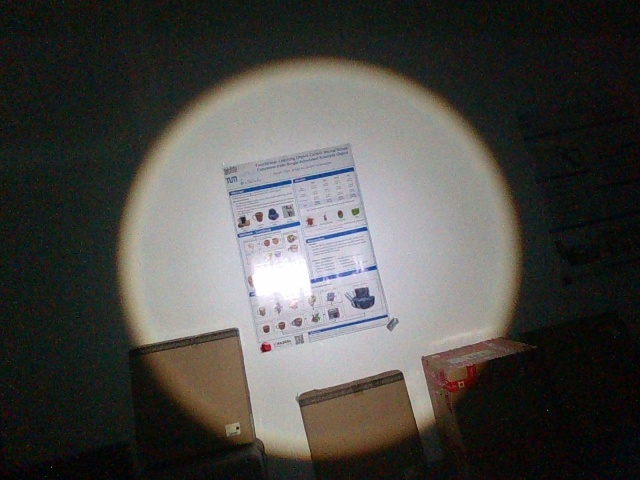} &
\Img{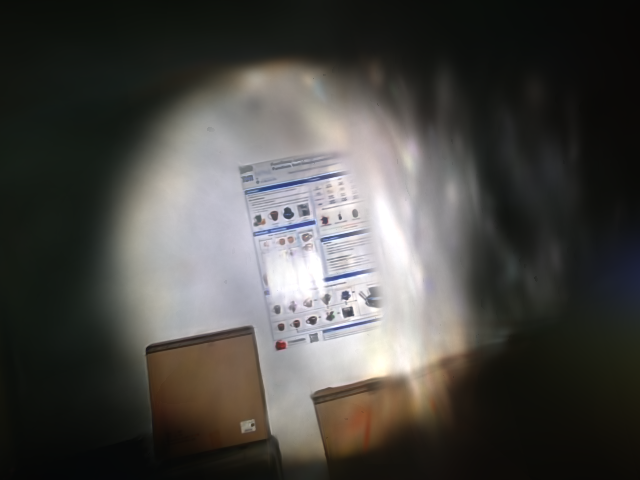} &
\Img{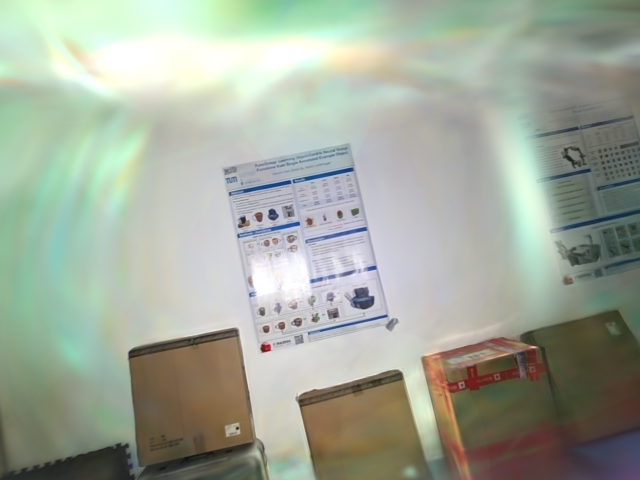} &
\Img{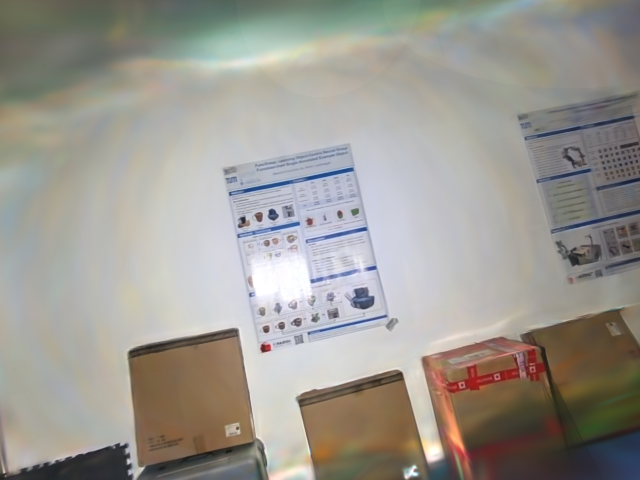} &
\Img{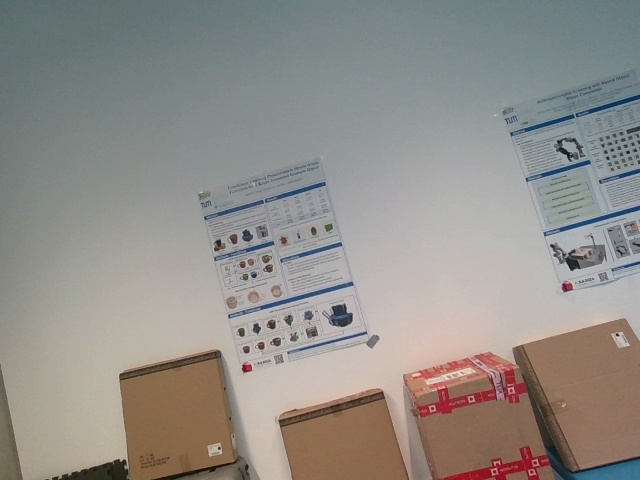} \\
\noalign{\vspace{1mm}}

\RowLabel{Self-col.} &
\Img{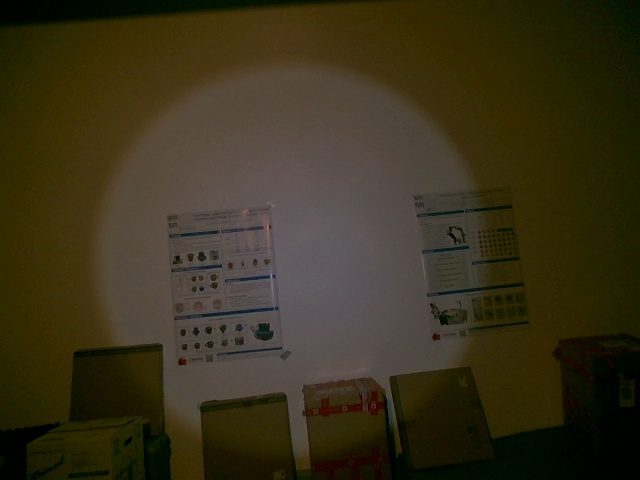} &
\Img{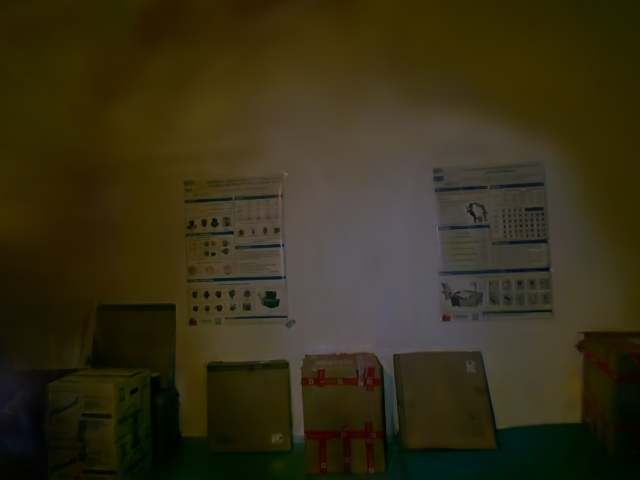} &
\Img{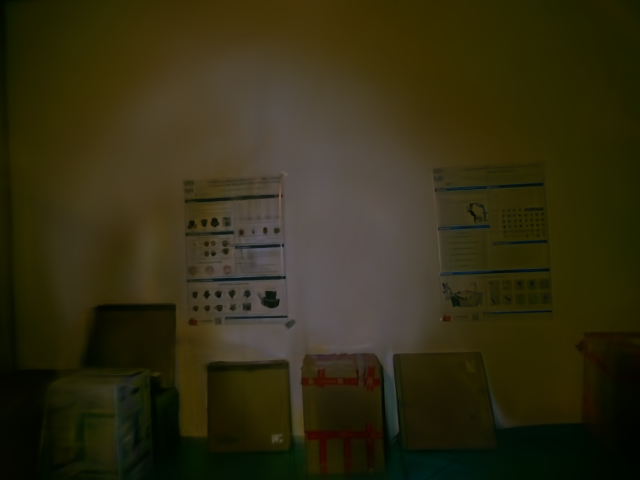} &
\Img{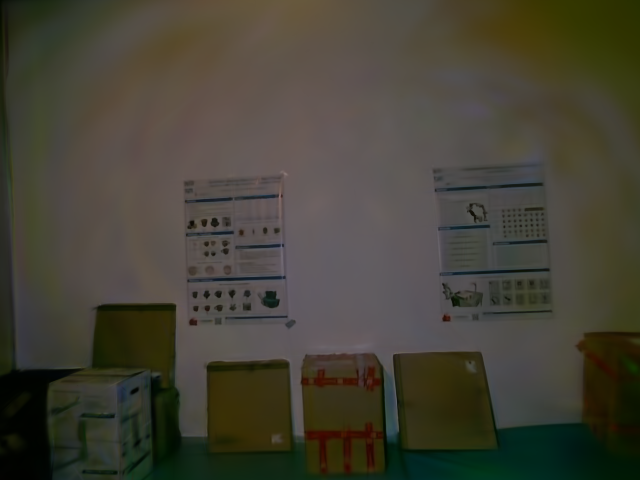} &
\Img{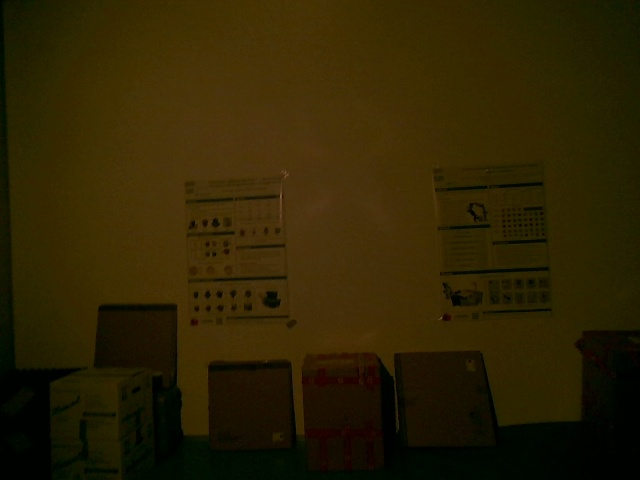} \\



\end{tabular}

\caption{Qualitative comparison on the synthetic (Uneven, Saturated) and self-collected (Self-col.) dataset. }
\label{fig:Qualitative evaluation result on synthetic and self-collected dataset}

\end{figure*}

We evaluate our method on four datasets, including a synthetic dataset, the DarkGS dataset, a self-collected real-world dataset, and two in-the-wild datasets (ANYmal and Vessel). Qualitatively, as shown in Figs.~\ref{fig:Qualitative evaluation result on synthetic and self-collected dataset} and~\ref{fig:Qualitative Evaluation on DarkGS Dataset}, Vanilla 3DGS struggles under local light sources because illumination is not explicitly modeled. DarkGS alleviates the strong over-exposure through its spotlight formulation, but exhibits residual spotlight artifacts, global illumination inconsistencies, color shifts, and reflection artifacts due to its simplified illumination model. In contrast, our method consistently produces more spatially coherent relighting results, effectively suppresses localized over-exposure, restores realistic global illumination, and better preserves fine structural and texture details across all datasets.

For datasets with available GT images (Synthetic and Self-collected), quantitative results in Table~\ref{tab:quantitative_results} demonstrate that our method achieves the best performance across PSNR, SSIM, and LPIPS, indicating better relighting fidelity and perceptual quality. On datasets without ground truth (DarkGS, ANYmal, and Vessel), visual comparisons further show that our approach more thoroughly mitigates spotlight artifacts and illumination inconsistencies while maintaining scene details, resulting in cleaner and more realistic relighting outcomes.

\begin{table}[t]
\centering
\caption{Quantitative Evaluation on Synthetic and Real-World Datasets.}
\label{tab:quantitative_results}
\setlength{\tabcolsep}{4pt}
\begin{tabular}{llccc}
\toprule
Dataset & Method & PSNR$\uparrow$ & SSIM$\uparrow$ & LPIPS$\downarrow$ \\
\midrule
\multirow{3}{*}{Synthetic Dataset}
& Vanilla 3DGS~\cite{kerbl20233d} & 6.04 & 0.391 & 0.424 \\
& DarkGS~\cite{zhang2024darkgs}       & 10.66 & 0.685 & 0.460 \\
& Ours         & \textbf{12.67} & \textbf{0.727} & \textbf{0.381} \\
\midrule
\multirow{3}{*}{Self-collected Dataset}
& Vanilla 3DGS~\cite{kerbl20233d} & 7.26 & 0.333 & 0.630 \\
& DarkGS~\cite{zhang2024darkgs}       & 11.09 & 0.749 & 0.604 \\
& Ours         & \textbf{17.46} & \textbf{0.804} & \textbf{0.513} \\
\bottomrule
\end{tabular}
\end{table}



\begin{figure*}[!tbp]
\centering

\setlength{\tabcolsep}{0.5mm} 

\newcommand{\imgw}{0.18\linewidth}
\newcommand{\labw}{1.3em}

\newcommand{\RowLabel}[1]{%
  \adjustbox{valign=c}{\rotatebox{90}{\small #1}}}

\newcommand{\Img}[1]{%
  \adjustbox{valign=c}{\includegraphics[width=\imgw]{#1}}}

\begin{tabular}{
@{} m{\labw} @{\hspace{1mm}}
>{\centering\arraybackslash}m{\imgw}
>{\centering\arraybackslash}m{\imgw}
>{\centering\arraybackslash}m{\imgw}
>{\centering\arraybackslash}m{\imgw}
>{\centering\arraybackslash}m{\imgw}
@{}}
& {\small Input}
& {\small Our Reconstructed}
& {\small Vanilla 3DGS}
& {\small DarkGS}
& {\small Ours}
\end{tabular}

\vspace{1mm} 

\begin{tabular}{
@{} >{\centering\arraybackslash}m{\labw} @{\hspace{1mm}}
c c c c c @{}
}

\RowLabel{DarkGS} &
\Img{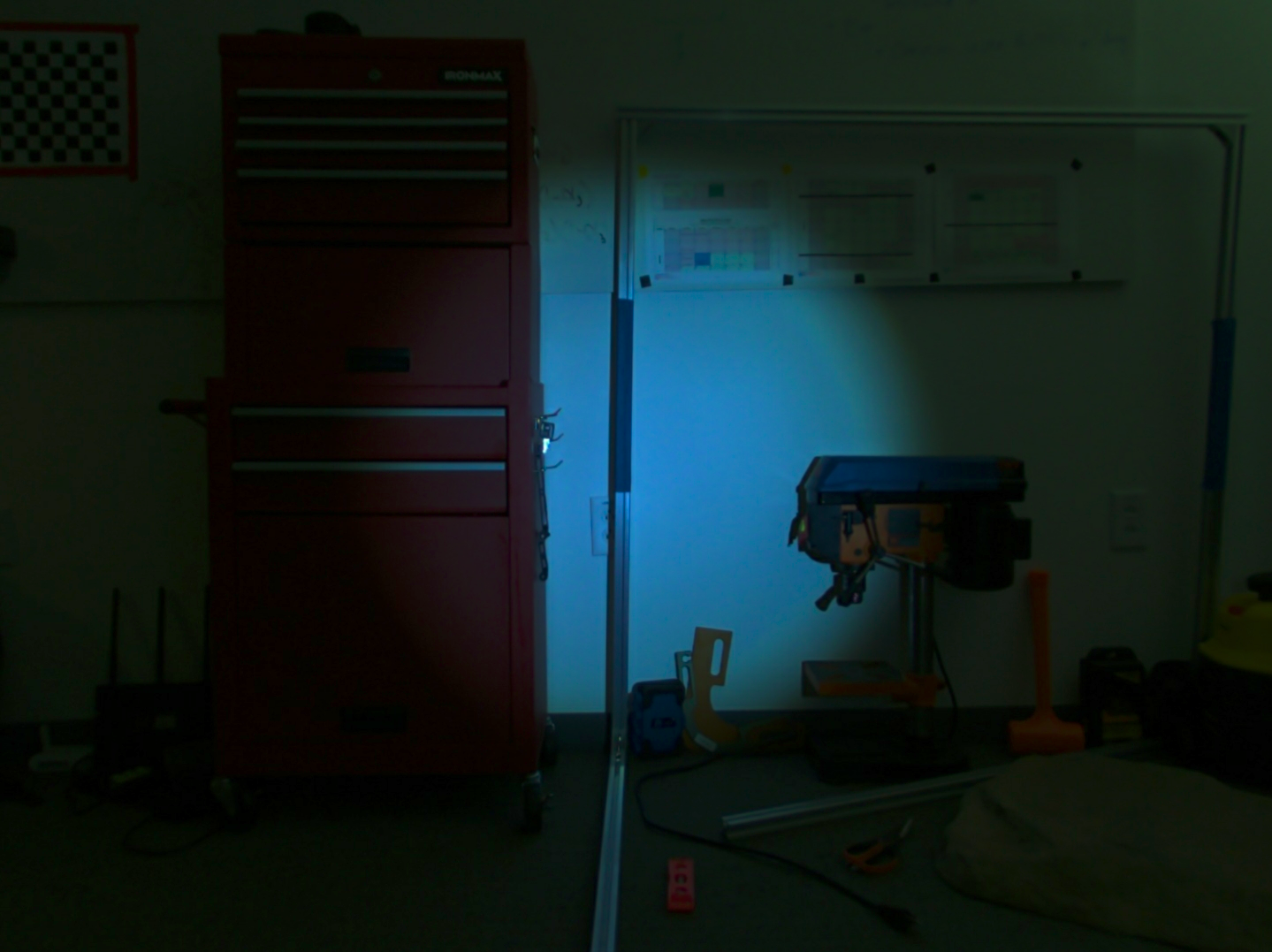} &
\Img{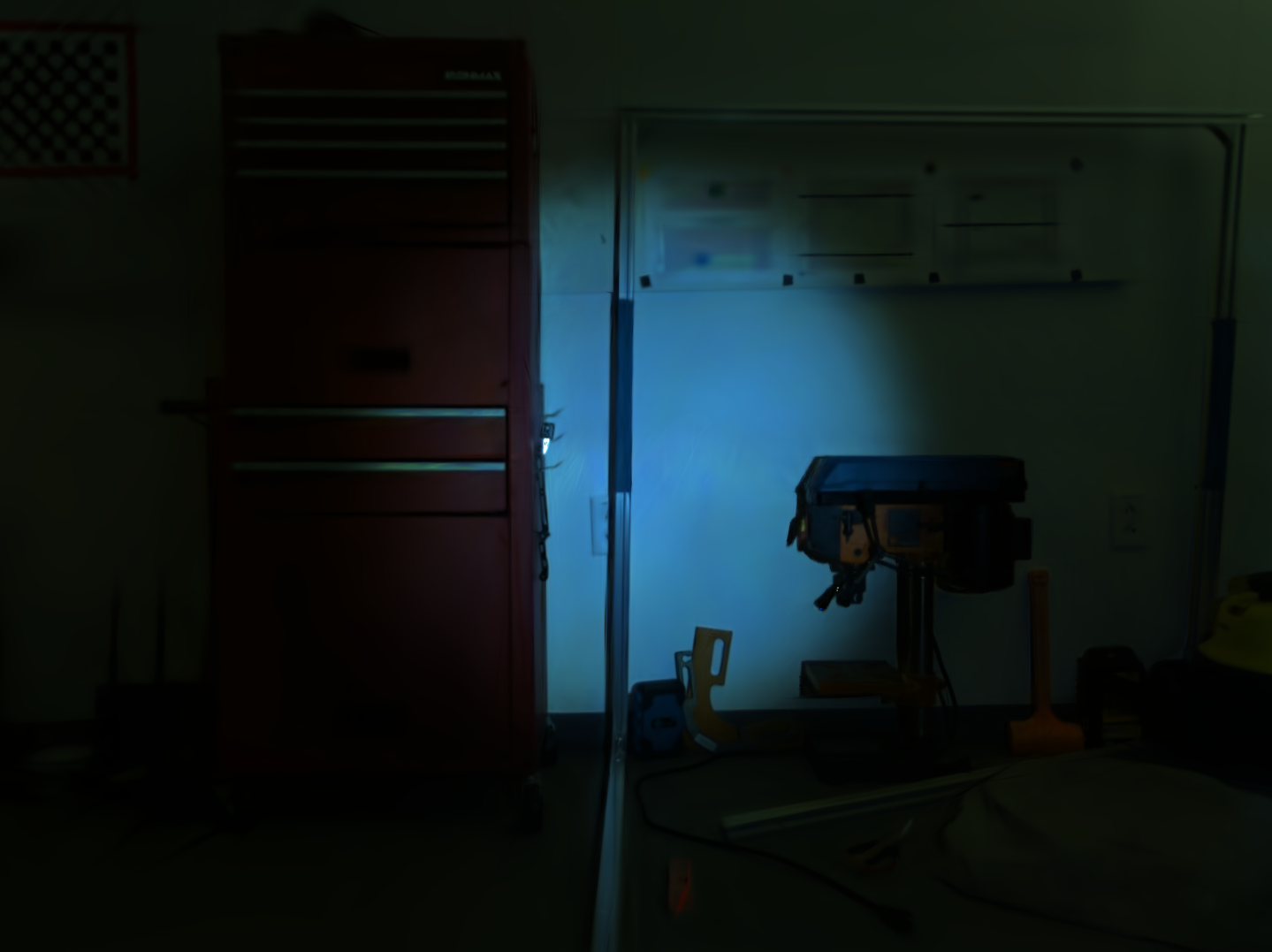} &
\Img{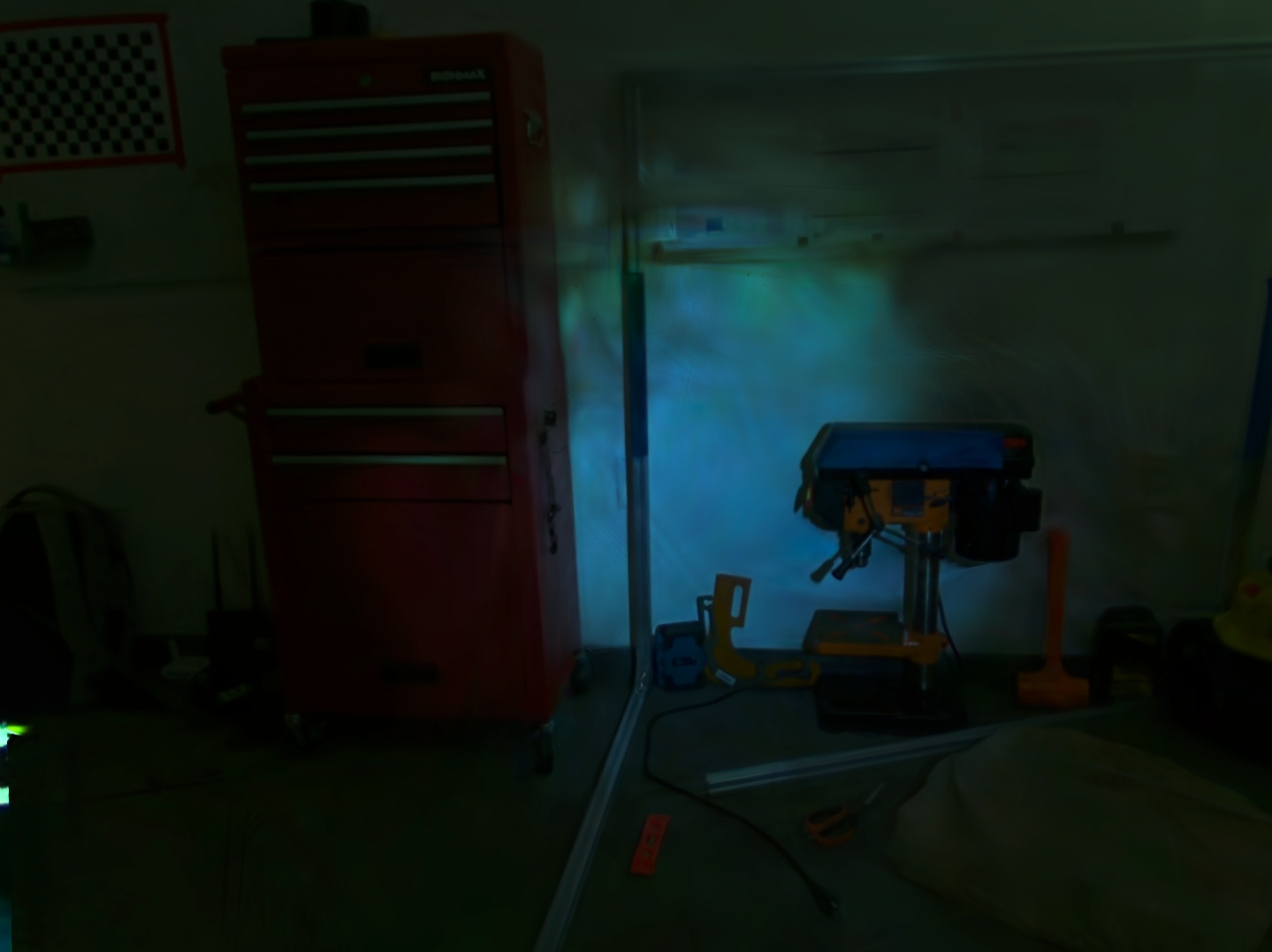} &
\Img{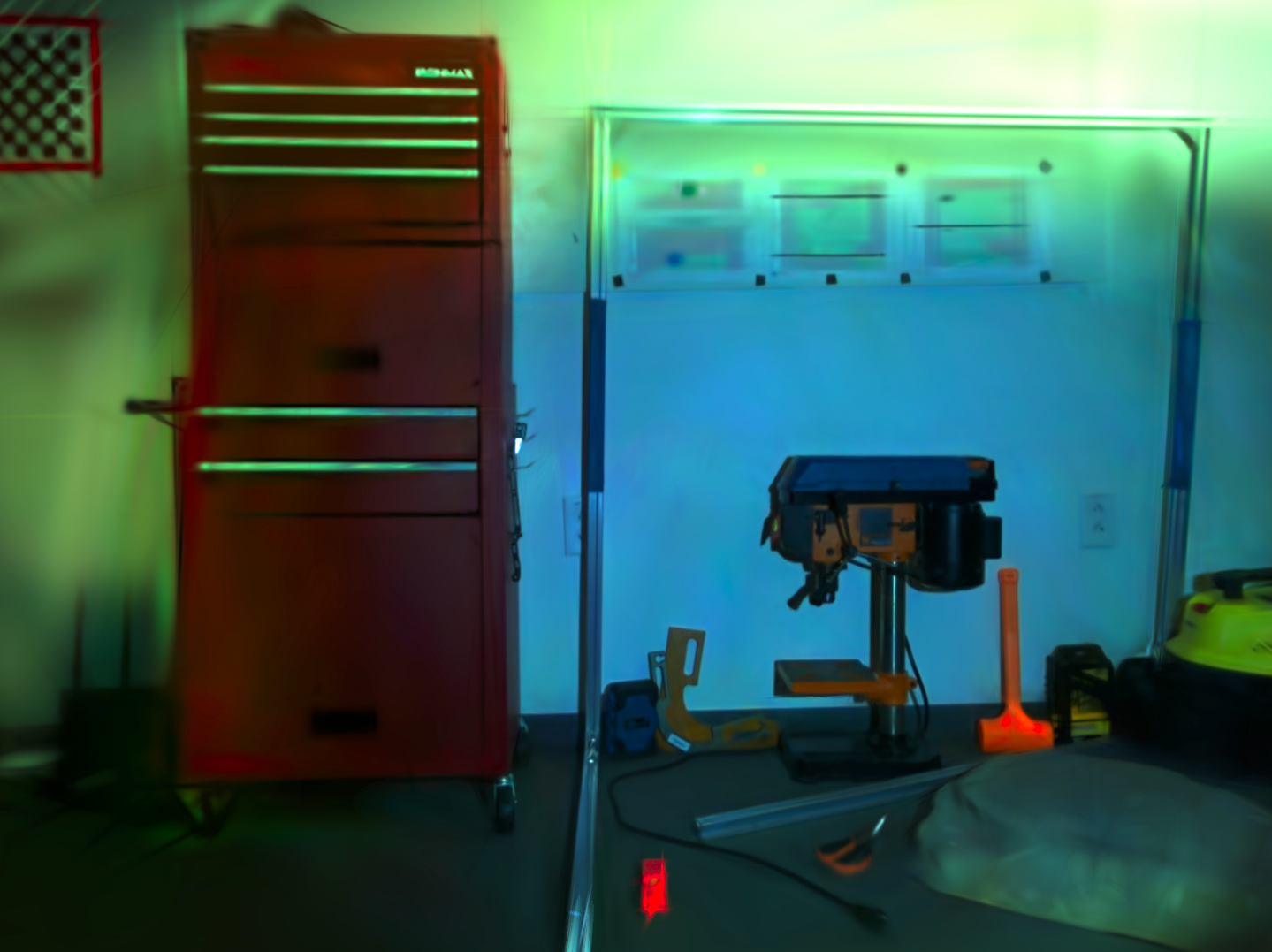} &
\Img{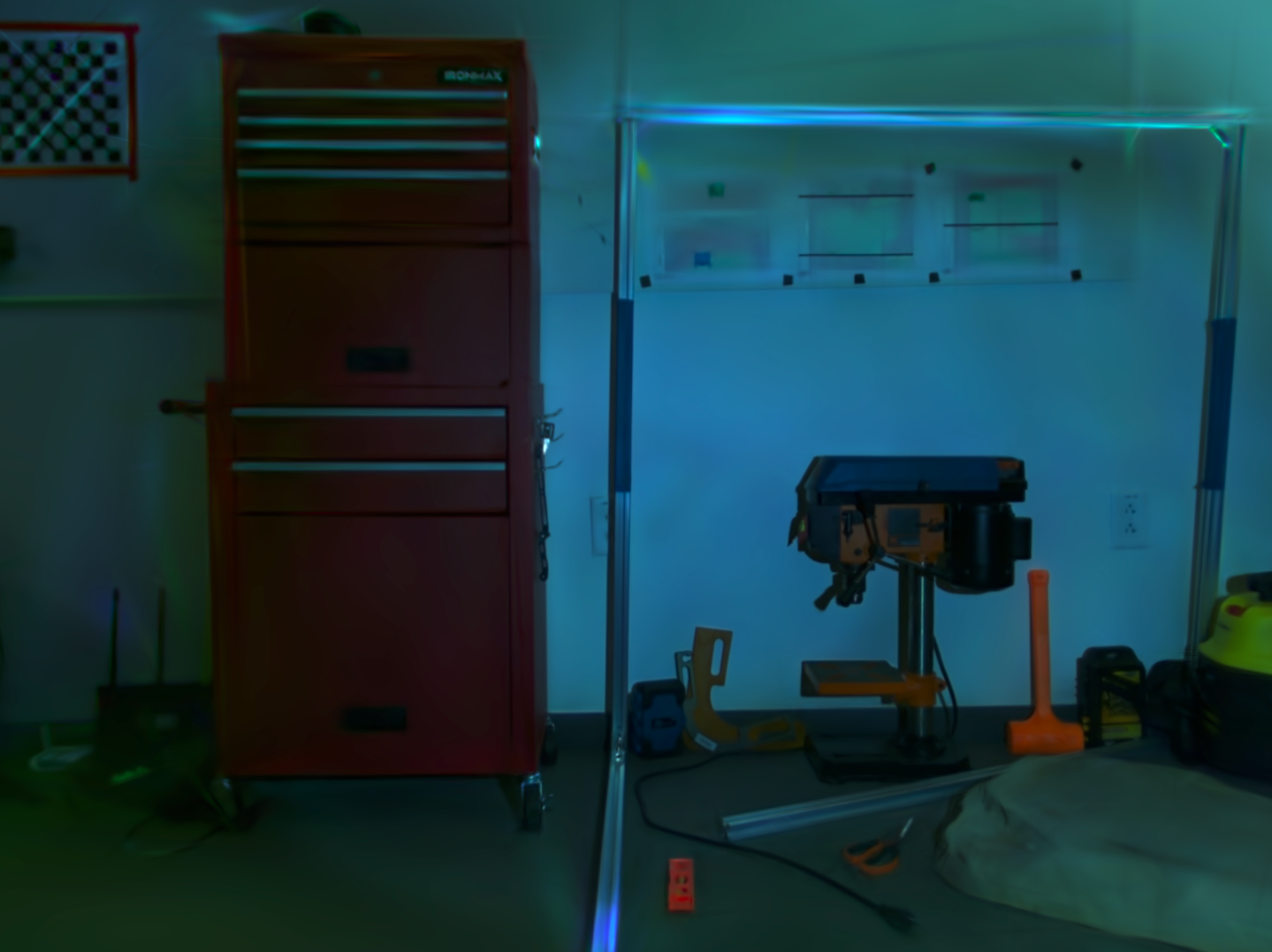} \\
\noalign{\vspace{1mm}}

\RowLabel{DarkGS} &
\Img{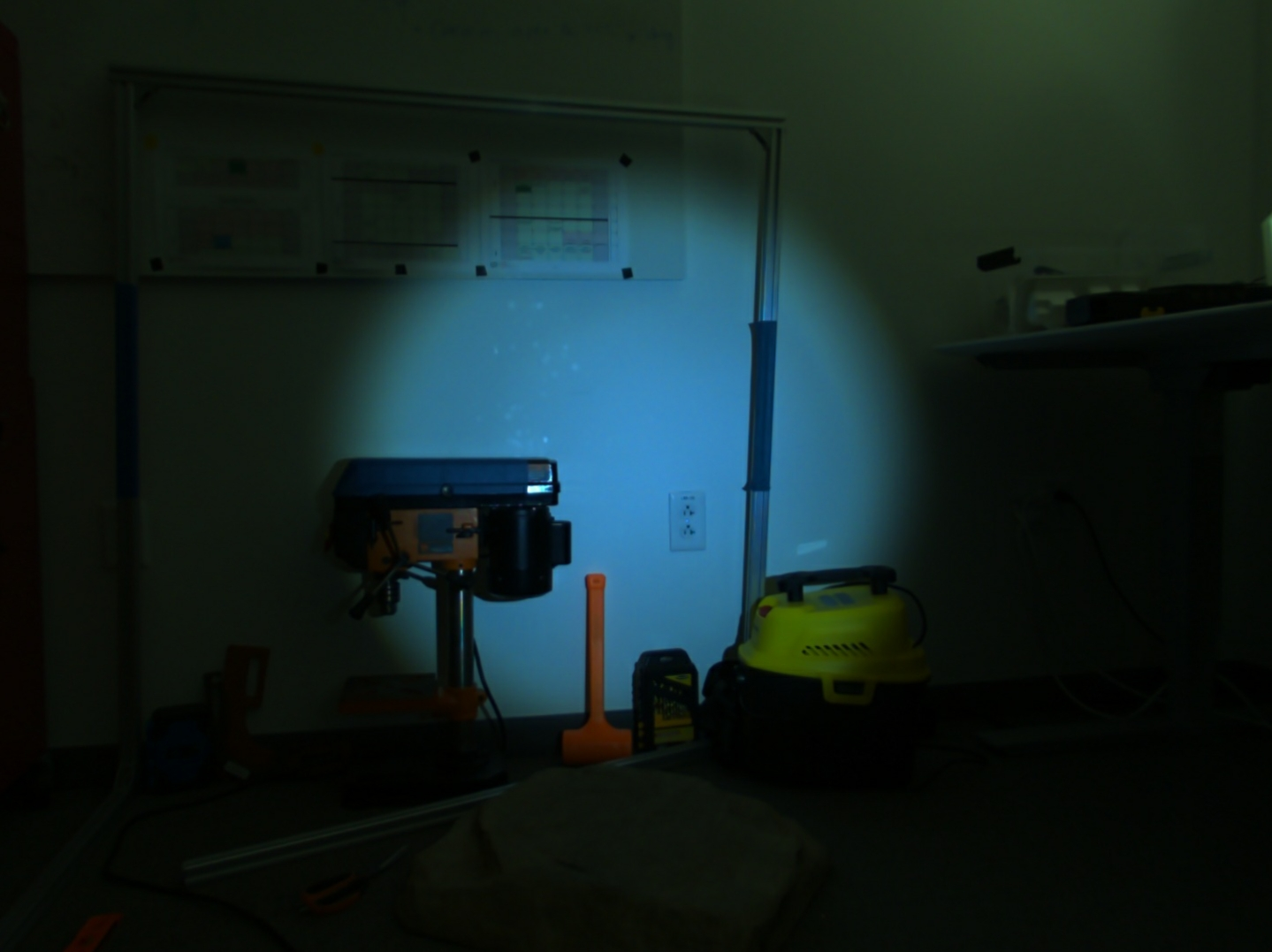} &
\Img{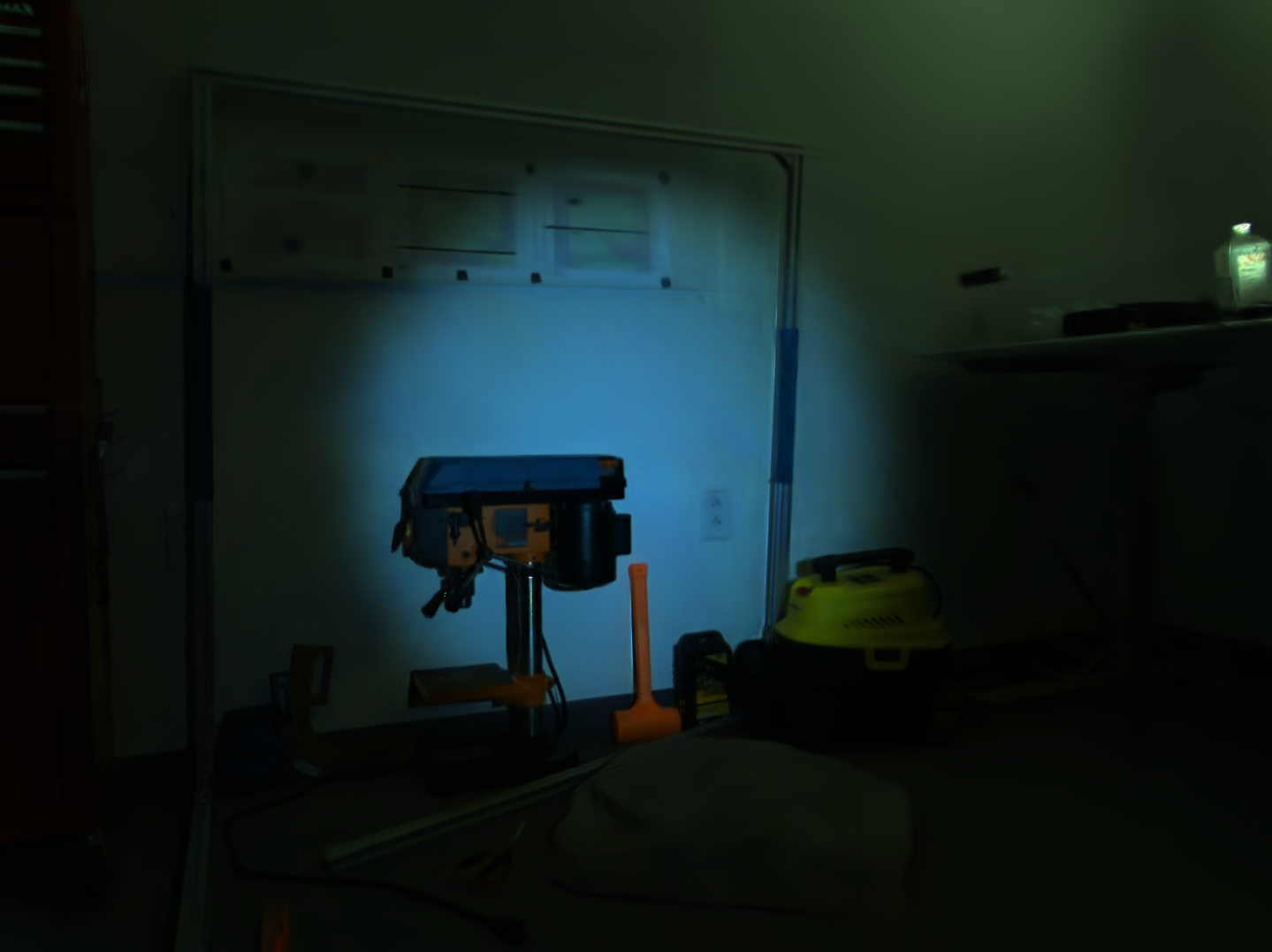} &
\Img{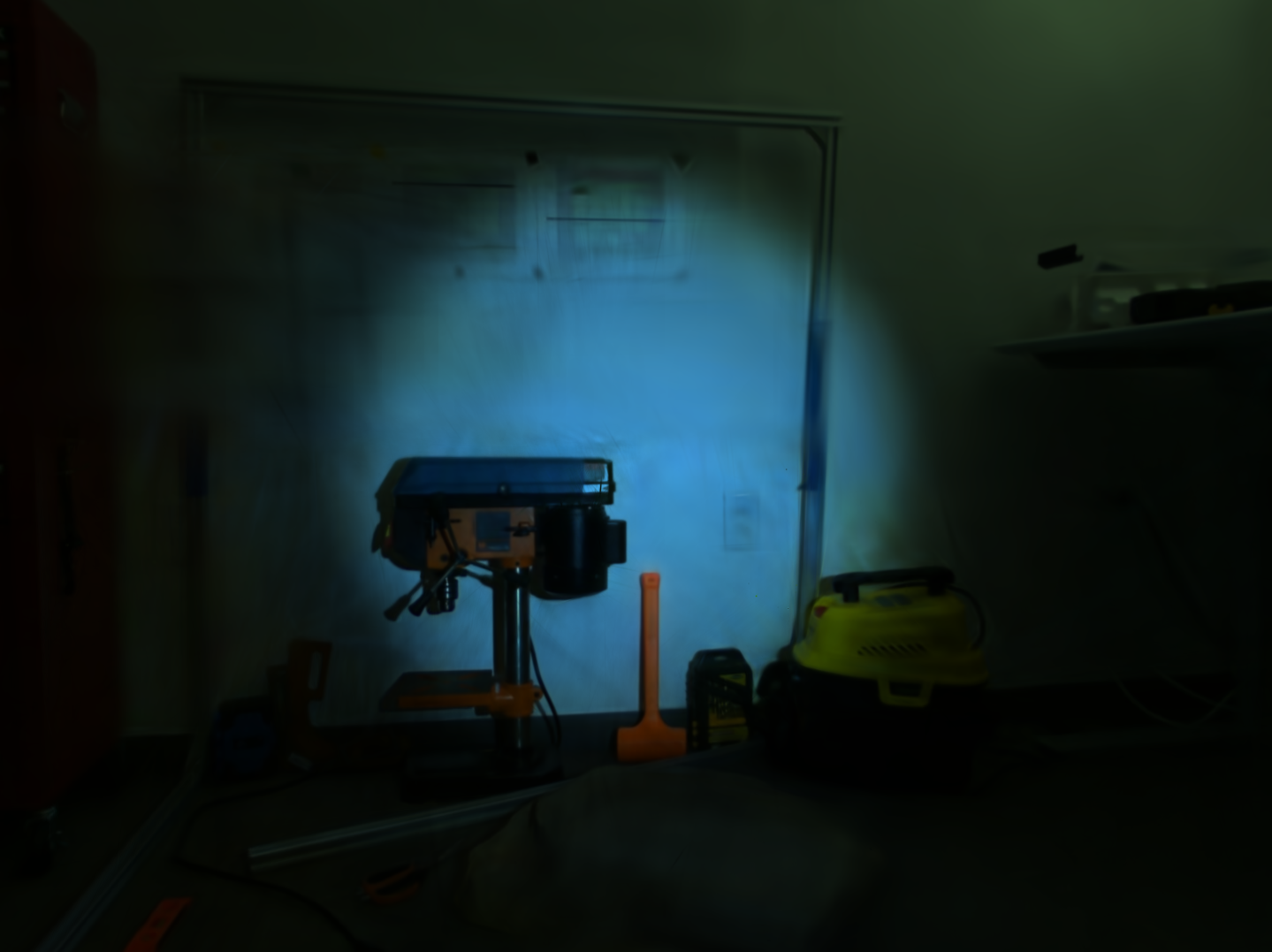} &
\Img{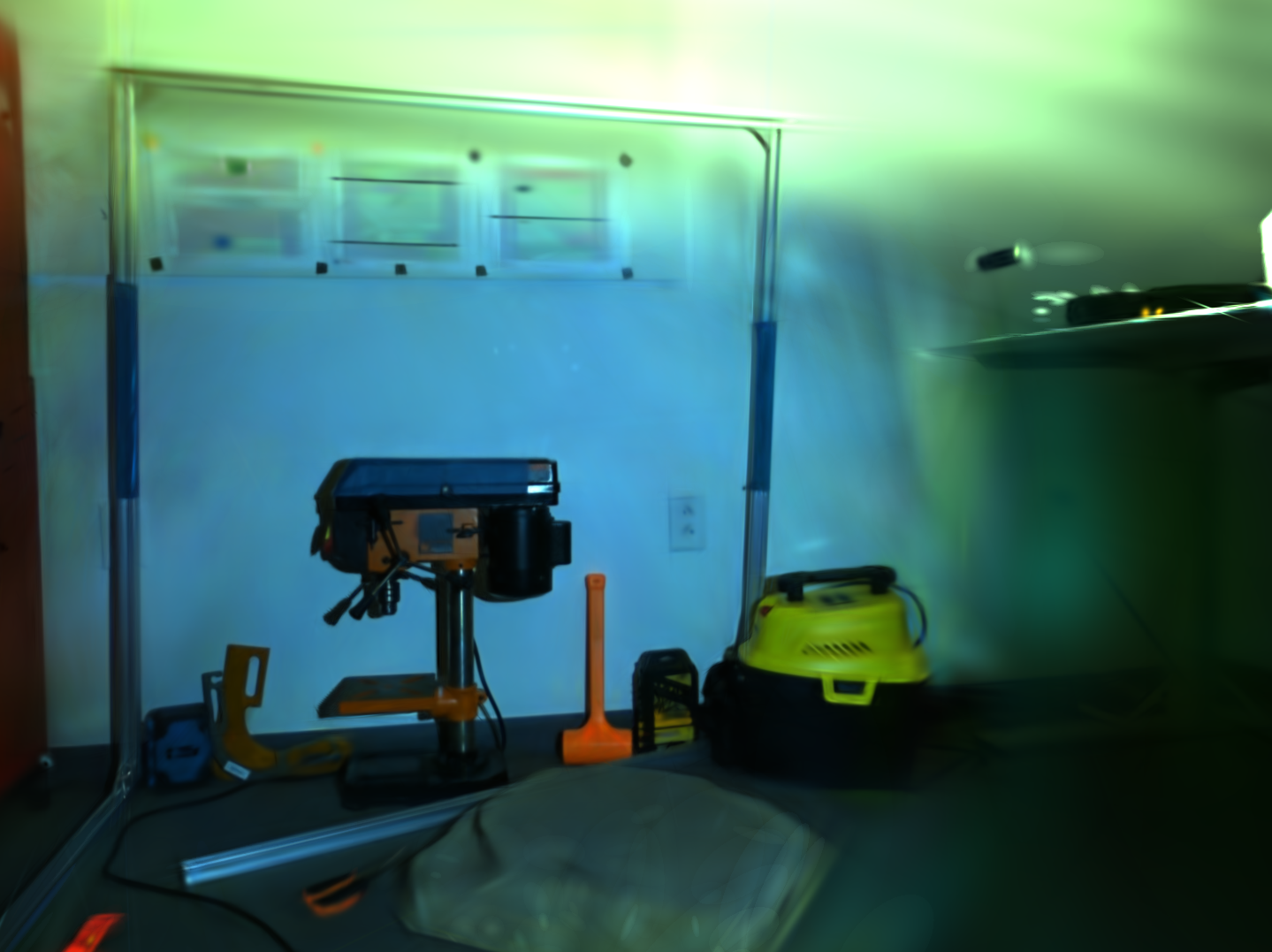} &
\Img{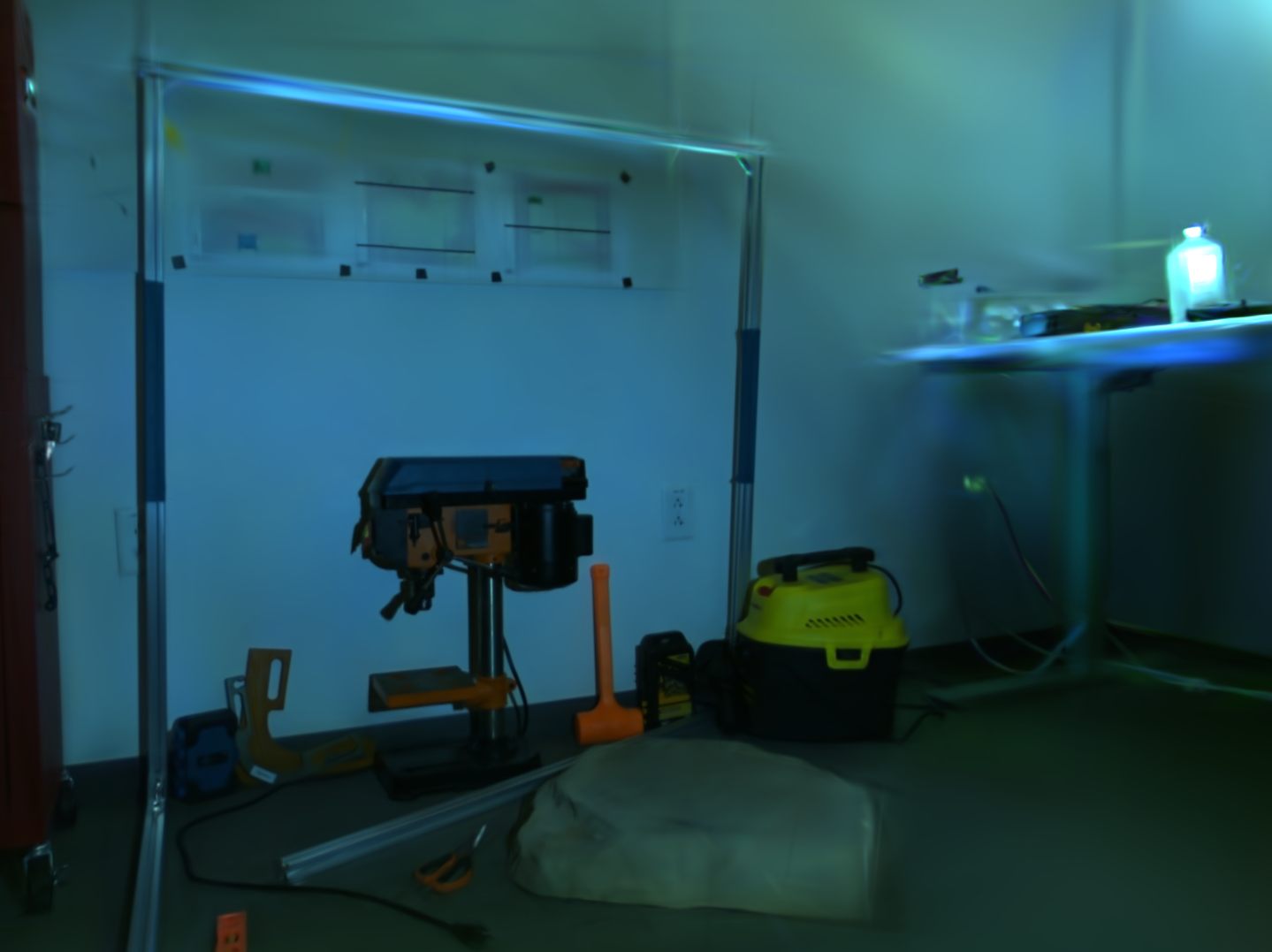} \\
\noalign{\vspace{1mm}}

\RowLabel{ANYmal} &
\Img{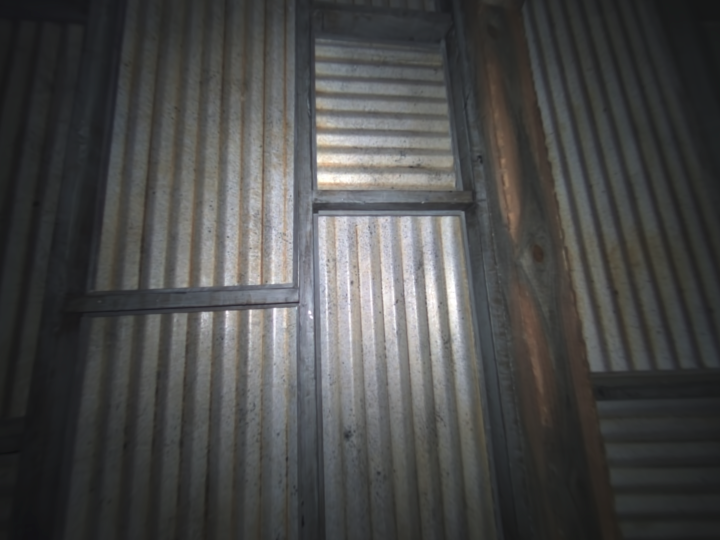} &
\Img{image_6.png} &
\Img{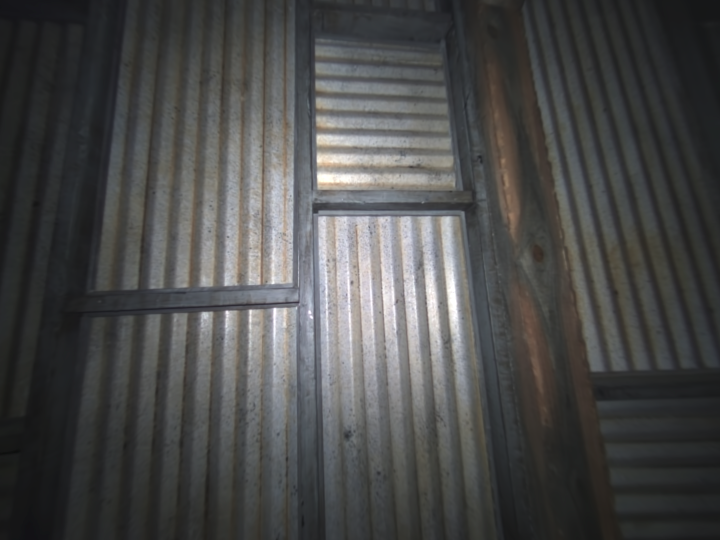} &
\Img{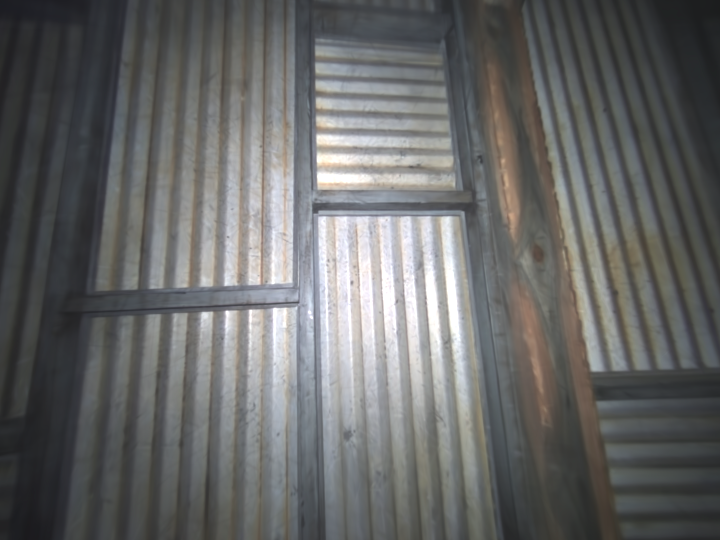} &
\Img{figures/real_world_dataset/Anymal/Ours/image_6.png} \\
\noalign{\vspace{1mm}}

\RowLabel{Vessel} &
\Img{figures/real_world_dataset/Vessel/Input/image_122.png} &
\Img{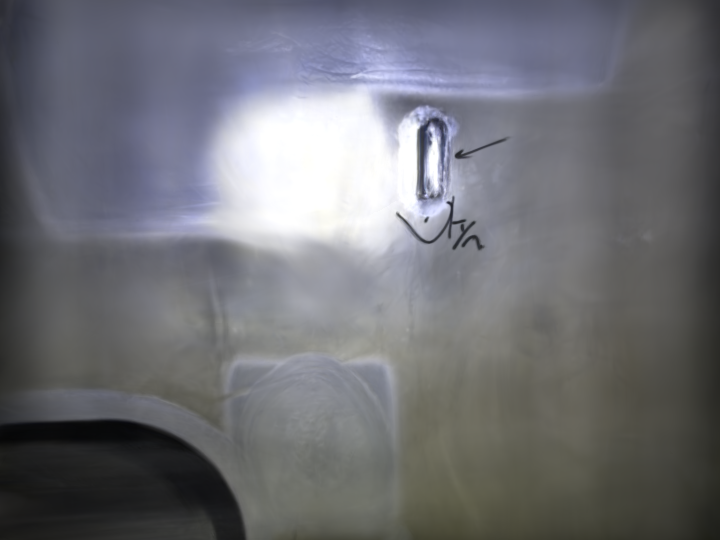} &
\Img{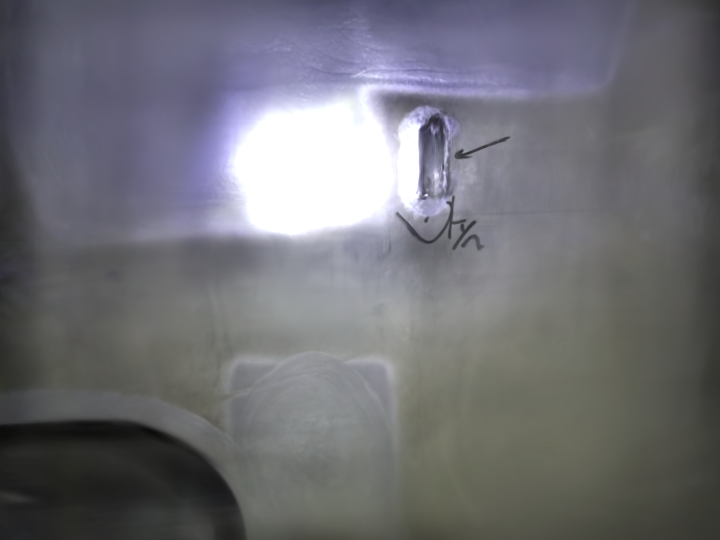} &
\Img{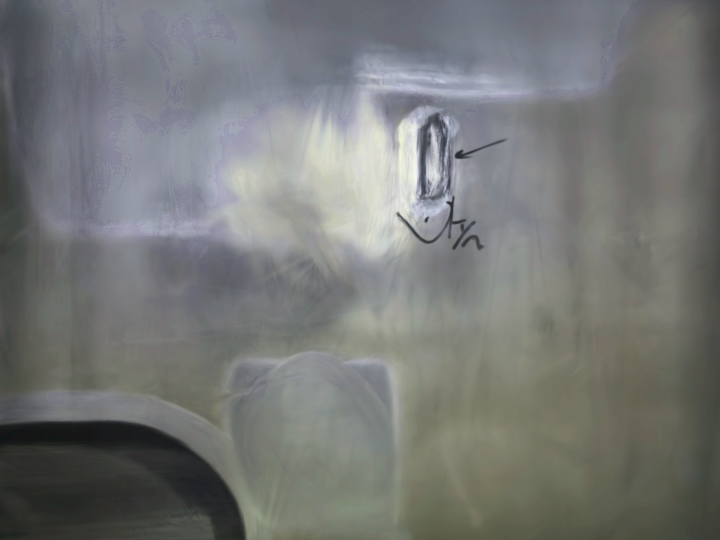} &
\Img{figures/real_world_dataset/Vessel/Ours/image_122.png} \\

\end{tabular}

\caption{Qualitative comparison on the DarkGS dataset~\cite{zhang2024darkgs} and real-world dataset in the wild~\cite{tranzatto2022cerberus, dharmadhikari2023autonomous}.} 
\label{fig:Qualitative Evaluation on DarkGS Dataset}

\end{figure*}




\subsubsection{Downstream Structure-from-Motion Evaluation}
\begin{figure}[t]
    \centering
    \includegraphics[width=0.819\linewidth]{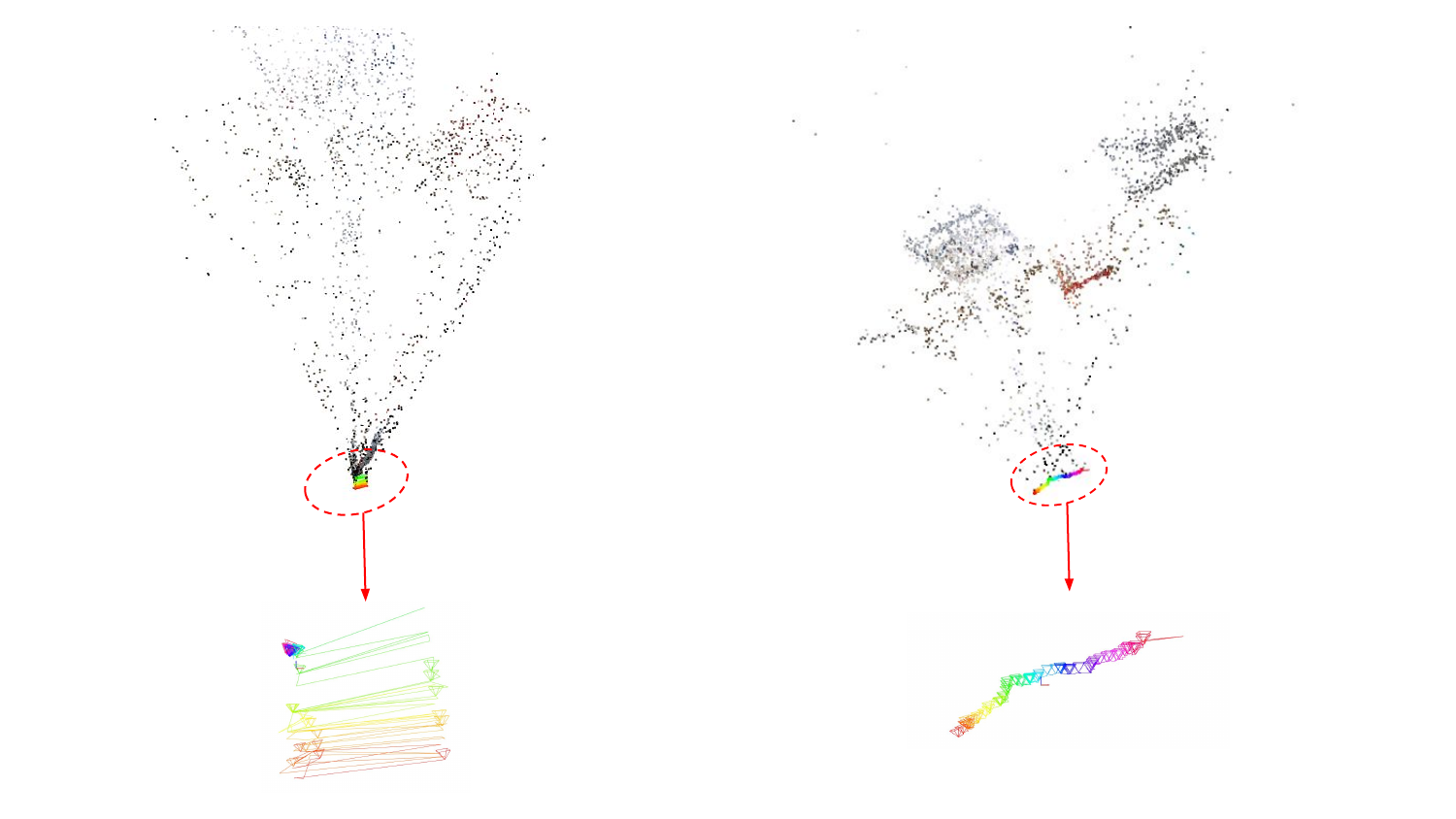}
    \caption{
    Downstream COLMAP reconstruction comparison. Left: Reconstruction from unevenly illuminated images. Right: Reconstruction from our relit images. Top: sparse point clouds with camera poses. Bottom: enlarged camera trajectories. Colors indicate temporal progression (blue $\rightarrow$ red). The relit images yield a smoother and physically plausible trajectory with more geometrically consistent point cloud.
    }
    \label{fig:downstream_colmap}
\end{figure}

To evaluate whether the proposed method improves real-world robotic perception beyond visual enhancement, we conduct a downstream structure-from-motion analysis using COLMAP~\cite{schoenberger2016sfm} on our self-collected dataset. We estimate camera trajectories and reconstruct sparse 3D models from both the original unevenly illuminated images and the relit images produced by our method. As shown in Fig.~\ref{fig:downstream_colmap}, given the approximately linear motion used during data collection, reconstructions from the raw inputs yield unstable and physically implausible camera trajectory. The corresponding sparse point clouds are also fragmented and structurally incoherent. In contrast, reconstructions from our relit images produce smoother and more spatially consistent camera trajectories that better align with the expected linear motion, and the point clouds more faithfully reflect the underlying scene geometry. Quantitative results in Table~\ref{tab:sfm_results} further support these observations. The mean reprojection error from relit images is substantially reduced, and the average number of observations per image increases. Together, these results demonstrate that the proposed method improves feature stability and geometric consistency, thereby enhancing the robustness of downstream 3D reconstruction and real-world robotic perception under challenging illumination conditions.

\begin{table}[t]
\centering
\caption{Quantitative Comparison on COLMAP reconstruction.}
\label{tab:sfm_results}
\begin{tabular}{lcccc}
\toprule
Method & Reprojection Error $\downarrow$ & Observations per Image $\uparrow$ \\
\midrule
Original (Uneven) & 1.164 px  & 480.75 \\
Vanilla 3DGS~\cite{kerbl20233d} & 0.733 px  & 290.76 \\
DarkGS~\cite{zhang2024darkgs}  & 0.611 px  & 881.78 \\
Ours         & \textbf{0.463 px} & \textbf{960.28} \\
\bottomrule
\end{tabular}
\end{table}

\subsubsection{Real-Time Performance}
We evaluated the runtime performance of our method on the Vessel dataset. We ran all experiments on a local machine equipped with an NVIDIA GeForce RTX 3070. Our method achieves an average inference time of $0.01\,\text{s}$ per frame ($101.4\,\text{FPS}$), compared to $0.004\,\text{s}$ ($228.1\,\text{FPS}$) for Vanilla GS and $0.003\,\text{s}$ ($369.6\,\text{FPS}$) for DarkGS. And memory usage is slightly higher for our method ($1.38\,\text{GB}$) compared to DarkGS ($1.26\,\text{GB}$). The lower FPS and higher memory usage of our approach is due to the additional computations for the learnable MLP-based BRDF and spherical harmonics-based low-frequency light. Despite this, the method remains well within practical limits, demonstrating that our method improves visual quality and relighting fidelity at a modest computational cost.

\subsection{Ablations}
\subsubsection{Effect of Ambient Light using Spherical Harmonics}
\begin{figure}[t]
    \centering
    \begin{subfigure}[b]{0.32\linewidth}
        \centering
        \includegraphics[width=\linewidth]{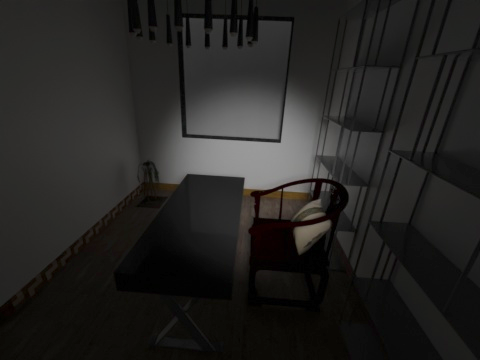}
        \caption{Input}
    \end{subfigure}
    \hfill
    \begin{subfigure}[b]{0.32\linewidth}
        \centering
        \includegraphics[width=\linewidth]{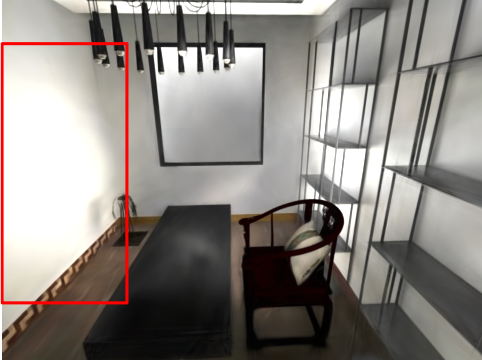}
        \caption{Const. ambient light}
    \end{subfigure}
    \hfill
    \begin{subfigure}[b]{0.32\linewidth}
        \centering
        \includegraphics[width=\linewidth]{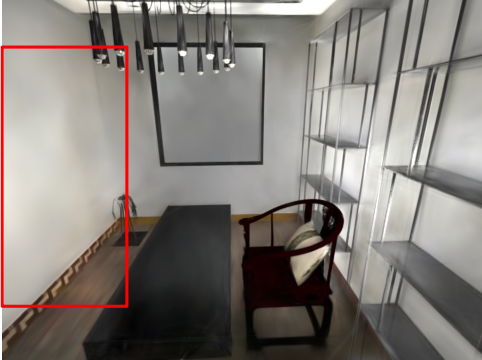}
        \caption{SH ambient light}
    \end{subfigure}
    \vfill
    \vspace{2mm}
    \begin{subfigure}[b]{0.32\linewidth}
        \centering
        \includegraphics[width=\linewidth]{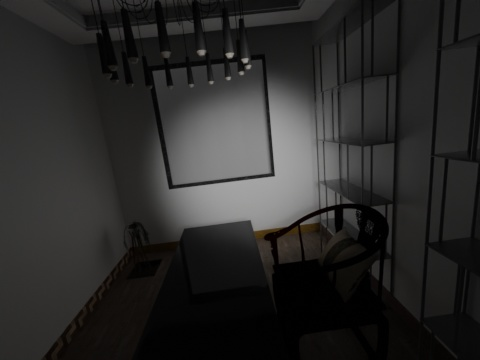}
        \caption{Input}
    \end{subfigure}
    \hfill
    \begin{subfigure}[b]{0.32\linewidth}
        \centering
        \includegraphics[width=\linewidth]{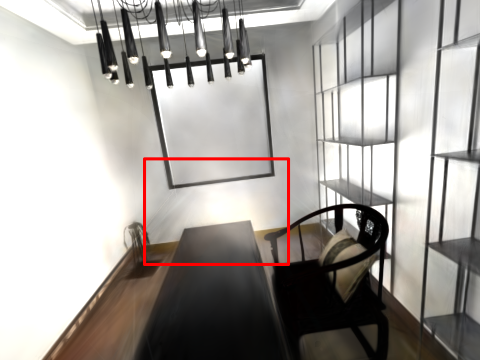}
        \caption{Lambertian BRDF}
    \end{subfigure}
    \hfill
    \begin{subfigure}[b]{0.32\linewidth}
        \centering
        \includegraphics[width=\linewidth]{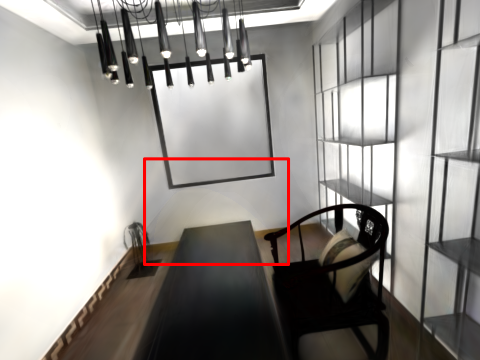}
        \caption{MLP-based BRDF}
    \end{subfigure}
    
    \caption{Qualitative results of the ablation study on SH-based ambient illumination and the MLP-based BRDF.}
    \label{fig:ablation}
\end{figure}

We analyze the impact of ambient light modeling in our relighting framework by comparing a constant ambient term with a spherical harmonics-based representation. As shown in Fig.~\ref{fig:ablation}(a)-(c), using a constant ambient term results in uneven environmental illumination, particularly on surfaces with directional shading (e.g., the left wall), as it cannot represent low-frequency directional lighting variations. In contrast, SH-based ambient lighting captures smooth directional dependencies, producing more uniform global illumination and improved structural consistency. Quantitatively, as reported in Table~\ref{tab:ablation_study}, SH-based modeling improves PSNR, SSIM and LPIPS over the baseline. These results demonstrate that incorporating SH basis functions enables more accurate approximation of global illumination compared to a constant bias term.

\begin{table}[t]
\centering
\caption{Ablation study on the effect of SH-based ambient light and MLP-based BRDF on the synthetic (Syn.) and real-world self-collected (Self-col.) datasets.}
\label{tab:ablation_study}
\setlength{\tabcolsep}{4pt}
\begin{tabular}{llccc}
\toprule
Dataset & Method & PSNR$\uparrow$ & SSIM$\uparrow$ & LPIPS$\downarrow$ \\
\midrule
\multirow{4}{*}{Syn. Dataset}
& DarkGS (Baseline)~\cite{zhang2024darkgs} & 10.66           & 0.685           & 0.460 \\
& + SH Ambient Light       & 12.62  & 0.724    &0.381 \\
& + MLP-based BRDF         & 10.85  & 0.721  &  0.403 \\
& + SH + MLP (Ours) & \textbf{12.67}  & \textbf{0.727}  &  \textbf{0.381} \\
\midrule
\multirow{4}{*}{Self-col. Dataset}
& DarkGS (Baseline)~\cite{zhang2024darkgs} & 11.09 & 0.749 & 0.604 \\
& + SH Ambient Light       & 15.96 & 0.768 & 0.524 \\
& + MLP-based BRDF         & 11.67 & 0.775 & 0.565 \\
& + SH + MLP (Ours) & \textbf{17.46} & \textbf{0.804} & \textbf{0.513} \\
\bottomrule
\end{tabular}
\end{table}

\subsubsection{Effect of MLP-based BRDF}
 We evaluate the effect of the MLP-based BRDF by comparing it with a Lambertian reflectance model. As shown in Fig.~\ref{fig:ablation}(d)-(f), the MLP-based BRDF produces sharper edges and more effectively suppresses residual spotlight artifacts, whereas the Lambertian model leaves visible over-exposed regions and fails to capture complex reflectance behavior. Quantitative results in Table~\ref{tab:ablation_study} show that although the SH-based ambient lighting model contributes the majority of the performance improvement, the MLP-based BRDF provides additional gains, particularly on the self-collected real-world dataset. Specifically, adding the MLP-based BRDF on top of SH-based ambient lighting improves PSNR from 15.96 dB to 17.46 dB on the self-collected dataset, compared to only a marginal improvement on the synthetic dataset (12.62 dB to 12.67 dB). We attribute the larger gain on the self-collected dataset to the more complex, non-Lambertian reflectance properties present in real-world scenes, where the learned MLP-based BRDF provides greater flexibility for modeling view-dependent appearance variations. In contrast, the synthetic dataset appears to be dominated by near-Lambertian surfaces, for which a Lambertian reflectance model is already sufficient, thereby limiting the additional benefits of the learned BRDF.

\subsubsection{Effect of Multi-view Rendering}

\begin{figure}[!tbp]
    \centering
    \begin{subfigure}[b]{0.32\linewidth}
        \centering
        \includegraphics[width=\linewidth]{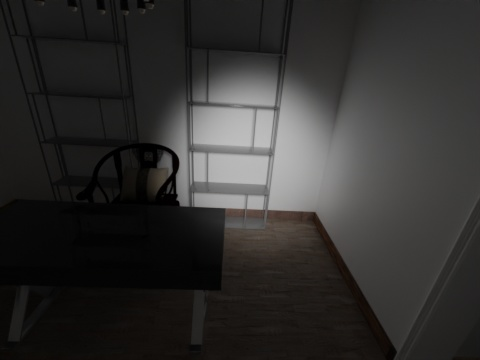}
    \end{subfigure}
    \hfill
    \begin{subfigure}[b]{0.32\linewidth}
        \centering
        \includegraphics[width=\linewidth]{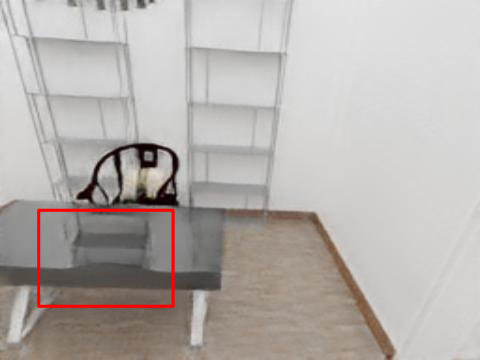}
    \end{subfigure}
    \hfill
    \begin{subfigure}[b]{0.32\linewidth}
        \centering
        \includegraphics[width=\linewidth]{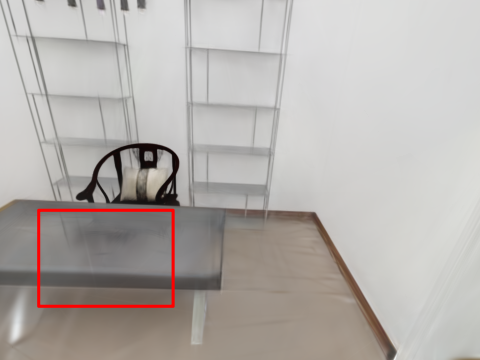}
    \end{subfigure} \\[1mm]

    \begin{subfigure}[b]{0.32\linewidth}
        \centering
        \includegraphics[width=\linewidth]{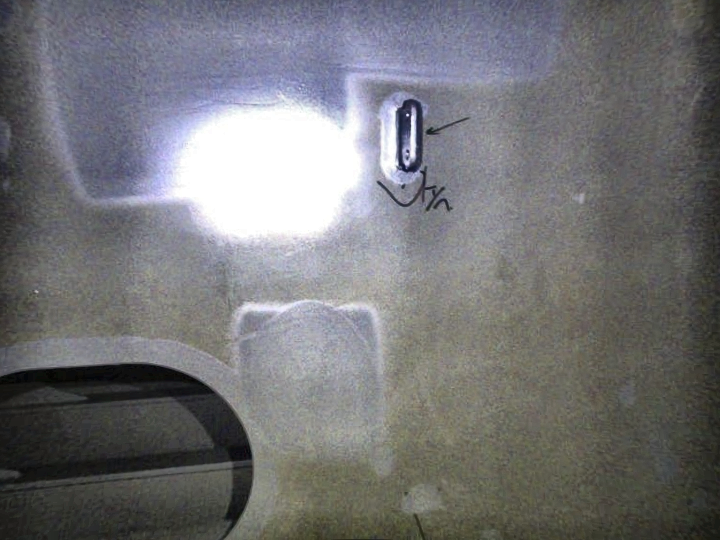}
        \caption{Input}
    \end{subfigure}
    \hfill
    \begin{subfigure}[b]{0.32\linewidth}
        \centering
        \includegraphics[width=\linewidth]{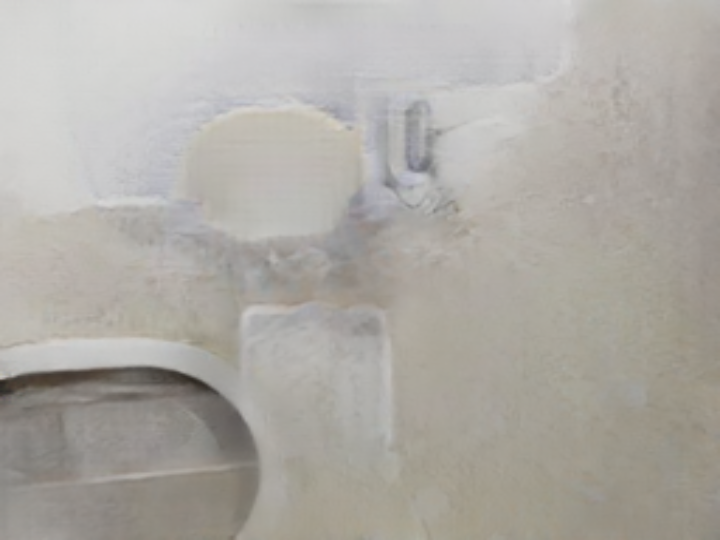}
        \caption{Cycle-GAN}
    \end{subfigure}
    \hfill
    \begin{subfigure}[b]{0.32\linewidth}
        \centering
        \includegraphics[width=\linewidth]{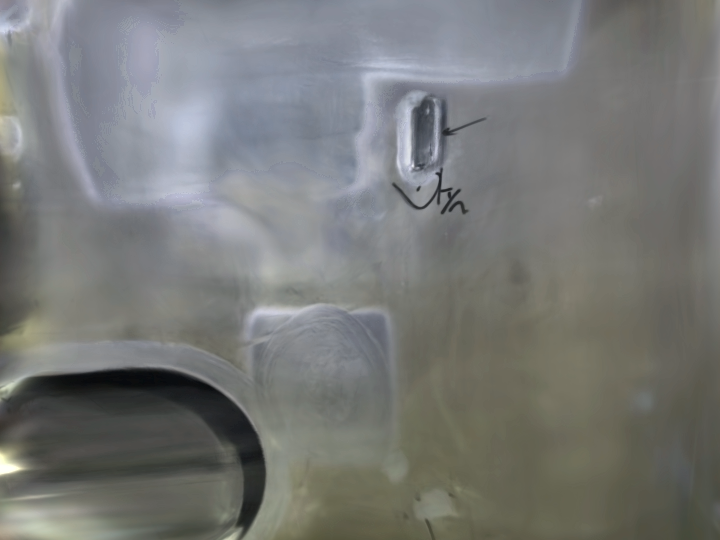}
        \caption{Multi-view}
    \end{subfigure}

    \caption{Effect of Multi-view Rendering}
    \label{fig:Effect of Multi-view Rendering}
\end{figure}

We evaluate the effect of multi-view training by comparing our method with a single-view baseline. The baseline employs Cycle-GAN~\cite{zhu2017unpaired} trained on an unpaired synthetic dataset to translate spotlight-illuminated images to an evenly lit domain. Although it improves appearance, it does not model the underlying 3D geometry, resulting in inconsistent shading across views. In contrast, our multi-view framework jointly optimizes geometry and illumination in 3D space, enforcing cross-view consistency during relighting. As shown in Fig.~\ref{fig:Effect of Multi-view Rendering}, the single-view baseline produces inconsistent illumination and structural artifacts, while our approach generates smoother and more stable results. Quantitative results in Table~\ref{tab:Effect of multi-view rendering on synthetic dataset.} demonstrate improvements in PSNR, SSIM, and temporal (Tmp.) metrics defined in Sec.~\ref{sec:eval-metric}, illustrating enhanced relighting performance and improved cross-view consistency across frames.

\begin{table}[!tbp]
    \centering
    \setlength{\tabcolsep}{4pt} 
    \caption{Effect of multi-view rendering on synthetic dataset.}
    \begin{tabular}{@{}lccccc@{}}
        \toprule
        Method                        & PSNR $\uparrow$            & SSIM $\uparrow$            &Tmp. PSNR $\uparrow$       &Tmp. SSIM $\uparrow$           \\
        \midrule
        Cycle-GAN~\cite{zhu2017unpaired}                     & 28.97           & 0.580           & 14.92          & 0.615\\
        Ours                          & \textbf{32.05}  & \textbf{0.887}  & \textbf{15.65}          &\textbf{0.664}\\
        \bottomrule
    \end{tabular}
    \label{tab:Effect of multi-view rendering on synthetic dataset.}
\end{table}

\section{Conclusion}


In this work, we present a relightable Gaussian Splatting framework to improve 3D reconstruction and rendering robustness under uneven onboard illumination. Building upon DarkGS, we introduce a more expressive reflectance and illumination formulation by incorporating a learnable MLP-based BRDF and spherical harmonics for ambient lighting within the Gaussian Splatting pipeline. This design enables spatially varying material modeling and produces more realistic lighting effects under onboard illumination. Extensive experiments demonstrate improved reconstruction quality and temporal stability over existing Gaussian Splatting methods, supporting more robust downstream robotic perception. Future work will explore integration with tasks such as SLAM and object detection.

\section*{Acknowledgment}
We would like to thank Yannick Burkhardt for his helpful suggestions on the real-world evaluation. We also acknowledge Dr. Cédric Le Gentil for his valuable feedback on the manuscript and his suggestions on the downstream application evaluation. We are also grateful to the anonymous reviewers for their constructive comments.

\bibliographystyle{IEEEtran}
\bibliography{bibliography}


\end{document}